\documentclass{article}

 \usepackage[main, final]{neurips_2026}

\usepackage[utf8]{inputenc} %
\usepackage[T1]{fontenc}    %
\usepackage{hyperref}       %
\usepackage{url}            %
\usepackage{booktabs}       %
\usepackage{amsfonts}       %
\usepackage{nicefrac}       %
\usepackage{microtype}      %
\usepackage{xcolor}         %

\title{Optimal Linear Regression Without a Variance}

\usepackage[utf8]{inputenc} %
\usepackage[T1]{fontenc}    %
\usepackage{hyperref}       %
\hypersetup{
    colorlinks,
    breaklinks,
    linkcolor = blue,
    citecolor = blue,
    urlcolor  = blue,
}
\usepackage{url}            %
\usepackage{amsfonts}       %
\usepackage{nicefrac}       %
\usepackage{microtype}      %
\usepackage{amsmath,bm}
\usepackage{amsthm}
\usepackage[algo2e, ruled, vlined]{algorithm2e}
\usepackage{algorithm}
\usepackage{bbm}      
\usepackage{mathtools}  \usepackage{matlab-prettifier}
\usepackage{amsopn}
\usepackage{amssymb}
\usepackage{graphicx}
\usepackage{xcolor}
\usepackage{footnote}
\usepackage{makecell}
\makesavenoteenv{tabular}
\makesavenoteenv{table}

\usepackage{booktabs,caption,threeparttable}

\usepackage{xcolor}
\usepackage{soul}
\usepackage{changepage}

\definecolor{Green}{rgb}{0.13, 0.65, 0.3}
\definecolor{Amber}{rgb}{0.3, 0.5, 1.0}
\usepackage[]{color-edits}
\usepackage{comment}

\usepackage{lipsum}
\usepackage{minitoc}
\usepackage{subcaption}

\newcommand{\tr}{\mathrm{Tr}}

\sethlcolor{gray}

\newcommand{\bbR}{\mathbb{R}}

\newcommand{\bi}{\begin{itemize}}
\newcommand{\ei}{\end{itemize}}

\newtheorem{theorem}{Theorem}
\newtheorem*{theorem*}{Theorem}
\newtheorem{lemma}{Lemma}

\newtheorem{corollary}{Corollary}

\newtheorem{assumption}{Assumption}

\newtheorem{proposition}{Proposition}
\newtheorem{remark}{Remark}

\newenvironment{sketch}{%
  \proof}{\endproof}

\usepackage{amssymb}%
\usepackage{pifont}%
\newcommand{\cmark}{\ding{51}}%
\newcommand{\xmark}{\ding{55}}%

\allowdisplaybreaks

\DeclareMathOperator*{\argmin}{\arg\!\min}

\usepackage{amsmath}
\usepackage{amssymb}
\usepackage{graphicx}
\usepackage{mathtools}
\usepackage{enumerate}
\usepackage{enumitem}
\usepackage{footnote}
\usepackage{float}
\usepackage{xspace}
\usepackage{multirow}
\usepackage{nicefrac}
\usepackage{wrapfig}
\usepackage{framed}
\usepackage{url}
\usepackage{tikz}
\usetikzlibrary{arrows,chains,matrix,positioning,scopes}

\newcommand{\calC}{{\mathcal{C}}}

\newcommand{\calY}{{\mathcal{Y}}}

\newcommand{\calD}{{\mathcal{D}}}

\newcommand{\calE}{{\mathcal{E}}}

\newcommand{\calF}{{\mathcal{F}}}

\newcommand{\field}[1]{\mathbb{#1}}

\newcommand{\E}{\field{E}}

\newcommand{\order}{\ensuremath{\mathcal{O}}}
\newcommand{\otil}{\ensuremath{\widetilde{\mathcal{O}}}}

\DeclareFontFamily{OMX}{MnSymbolE}{}
\DeclareFontShape{OMX}{MnSymbolE}{m}{n}{
    <-6>  MnSymbolE5
   <6-7>  MnSymbolE6
   <7-8>  MnSymbolE7
   <8-9>  MnSymbolE8
   <9-10> MnSymbolE9
  <10-12> MnSymbolE10
  <12->   MnSymbolE12}{}
\DeclareSymbolFont{mnlargesymbols}{OMX}{MnSymbolE}{m}{n}
\SetSymbolFont{mnlargesymbols}{bold}{OMX}{MnSymbolE}{b}{n}
\DeclareMathDelimiter{\llangle}{\mathopen}{mnlargesymbols}{'164}{mnlargesymbols}{'164}
\DeclareMathDelimiter{\rrangle}{\mathclose}{mnlargesymbols}{'171}{mnlargesymbols}{'171}

\usepackage{prettyref}
\newcommand{\pref}[1]{\prettyref{#1}}

\newcommand{\savehyperref}[2]{\texorpdfstring{\hyperref[#1]{#2}}{#2}}
\newrefformat{eq}{\savehyperref{#1}{Eq.~\eqref{#1}}}
\newrefformat{eqn}{\savehyperref{#1}{Equation~\eqref{#1}}}
\newrefformat{lem}{\savehyperref{#1}{Lemma~\ref*{#1}}}
\newrefformat{def}{\savehyperref{#1}{Definition~\ref*{#1}}}
\newrefformat{line}{\savehyperref{#1}{line~\ref*{#1}}}
\newrefformat{thm}{\savehyperref{#1}{Theorem~\ref*{#1}}}
\newrefformat{corr}{\savehyperref{#1}{Corollary~\ref*{#1}}}
\newrefformat{cor}{\savehyperref{#1}{Corollary~\ref*{#1}}}
\newrefformat{col}{\savehyperref{#1}{Corollary~\ref*{#1}}}
\newrefformat{sec}{\savehyperref{#1}{Section~\ref*{#1}}}
\newrefformat{app}{\savehyperref{#1}{Appendix~\ref*{#1}}}
\newrefformat{assum}{\savehyperref{#1}{Assumption~\ref*{#1}}}
\newrefformat{ex}{\savehyperref{#1}{Example~\ref*{#1}}}
\newrefformat{fig}{\savehyperref{#1}{Figure~\ref*{#1}}}
\newrefformat{tab}{\savehyperref{#1}{Table~\ref*{#1}}}
\newrefformat{medpe}{\savehyperref{#1}{Algorithm~\ref*{#1}}}
\newrefformat{alg}{\savehyperref{#1}{Algorithm~\ref*{#1}}}
\newrefformat{rem}{\savehyperref{#1}{Remark~\ref*{#1}}}
\newrefformat{conj}{\savehyperref{#1}{Conjecture~\ref*{#1}}}
\newrefformat{prop}{\savehyperref{#1}{Proposition~\ref*{#1}}}
\newrefformat{proto}{\savehyperref{#1}{Protocol~\ref*{#1}}}
\newrefformat{prob}{\savehyperref{#1}{Problem~\ref*{#1}}}
\newrefformat{claim}{\savehyperref{#1}{Claim~\ref*{#1}}}
\newrefformat{que}{\savehyperref{#1}{Question~\ref*{#1}}}
\newrefformat{obs}{\savehyperref{#1}{Observation~\ref*{#1}}}

\author{%
  Artin Tajdini \\
  University of Washington \\
  \texttt{artin@cs.washington.edu} \And
  Jasper C.H. Lee \\
  University of California, Davis \\
  \texttt{jasperlee@ucdavis.edu} \\
  \And
  Jonathan Scarlett \\
   National University of Singapore \\
  \texttt{scarlett@comp.nus.edu.sg} \\
  \And
  Kevin Jamieson \\
  University of Washington \\
  \texttt{jamieson@cs.washington.edu} \\
}

\begin{document}

\maketitle

\begin{abstract}
    We study high-probability linear regression in the infinite variance regime, where
the regression noise has only a finite conditional $\alpha$-moment $\nu$ for some
unknown $\alpha\in(1,2)$, and the covariates may only have finite moments up to a given order (e.g., 4th moment).
We give
a black-box reduction that converts finite-variance regression procedures into
weak-moment procedures.  The reduction is based on empirical residual pruning
and is agnostic to the moment exponent, the noise scale, the covariance, and
the design moment. 
Under a finite fourth-moment condition on the covariates, any finite-variance
subroutine attaining the sub-Gaussian rate yields the following when the number of samples $n$ is above a suitably defined threshold: Without knowing $\alpha$ or $\nu$, we have with probability at least $1-\delta$ that
$
    \|\widehat\theta-\theta^*\|_\Sigma
    \lesssim_\alpha
    \nu^{1/\alpha}
    \big(\frac{d+\log(1/\delta)}{n}\big)^{\frac{\alpha-1}{\alpha}},
$
where $d$ is the dimension, $\Sigma$ is the covariate covariance matrix, and
$\|v\|_\Sigma:=(v^\top\Sigma v)^{1/2}$.
We further
prove a matching minimax lower bound that holds under the same conditions. 
The reduction is computationally efficient whenever the finite-variance
subroutine is.  We also obtain a distribution-free prediction guarantee under a
bounded raw $\alpha$-moment assumption on the response, and give
an analogous
$\alpha$-agnostic reduction for mean estimation.

\end{abstract}

\section{Introduction}

Statistical estimation under heavy-tailed distributions has received sustained
attention since Catoni's work on robust mean and variance estimation
\citep{Catoni2012}.  A central theme in this line of work is that finite moment
assumptions exhibit poor concentration of empirical averages, so
high-probability guarantees require estimators designed for heavier tails.  The
mean-estimation problem is now well understood under finite (co)variance, both in
one dimension and in high dimensions
\citep{LugosiMendelson2019,pmlr-v99-cherapanamjeri19b,hopkins2020mean,lee2022optimal,lee2022optimalhigh}, and more recent work gives optimal
rates when the variance is infinite \citep{Bub13a,devroye2015subgaussianmeanestimators, cherapanamjeri2022optimal,
whitehouse2024mean,lee2025all}.  
In this paper, we study the corresponding question for linear regression.

Linear regression adds the geometry of the covariates to the heavy-tailed
estimation problem, and robust procedures typically require
some control of the covariate distribution.  Sharp finite-variance guarantees
are known under conditions such as finite fourth moments, hypercontractivity,
or small-ball compatibility; see \pref{tab:comparison} and the related work
below.  Concretely, with $d$ parameters, $n$ data points, and noise variance $\sigma^2$, the prediction-norm
rate is
\[
    \sigma
    \left(\frac{d+\log(1/\delta)}{n}\right)^{1/2}
\]
up to constants.\footnote{This result assumes positive-definite covariance.}
Importantly, ordinary least squares is insufficient when only finite variance noise is assumed, and prior works have developed a number of robust regression procedures in such settings.
However, the ``below-variance'' regime, where the regression error has only an $\alpha$-moment
for $1<\alpha<2$, is less developed.\footnote{To our knowledge, the best known lower bound for $\Sigma = I$ is from \cite{sun2020adaptive}, and equals $\left(\frac{\log (\nicefrac{1}{\delta})}{n}\right)^\frac{\alpha - 1}{\alpha}$ as the dimension-dependent terms cancel out; this is equal to mean estimation rate in $\bbR$.}  We also refer to this setting as \emph{weak-moment regression}.

Formally, our modeling assumptions are stated as follows.

\begin{assumption}[Linear model and error moment]
\label{assum:error-moment}
There is a parameter $\theta^*\in\bbR^d$ such that
\[
    y=x^\top\theta^*+\varepsilon,
    \qquad
    \E[\varepsilon\mid x]=0,
    \qquad
    \E[|\varepsilon|^\alpha\mid x]\le \nu,
\]
for some unknown $1<\alpha\le2$ and $\nu>0$.  The conditional distribution of $\varepsilon$ (given $x$) may depend on $x$, i.e., the noise need not be ``purely additive''.
\end{assumption}

Given $n$ i.i.d.~samples, a natural pessimistic baseline is to reduce the problem coordinatewise to one-dimensional weak-moment estimation (using scalar robust mean estimators such as \citep{Bub13a,devroye2015subgaussianmeanestimators}), which gives the rate
\(\otil\Big(\nu^{1/\alpha}\sqrt d
\big(\frac{\log(1/\delta)}{n}\big)^{(\alpha-1)/\alpha}\Big)\)
and pays a multiplicative $\sqrt d$ penalty.  Under sub-Gaussian covariates,
Huber-loss methods already achieve sharp high-probability guarantees
\citep{sun2020adaptive}, improving the $d$ dependence to $d^\frac{\alpha - 1}{\alpha}$. %
The question here is whether one can remove the
coordinatewise loss without such light-tailed design assumptions.
The central question of this paper is:

\begin{center}
\emph{
Can one improve on the coordinatewise rate
\(\otil\Big(\nu^{1/\alpha}\sqrt d
\big(\frac{\log(1/\delta)}{n}\big)^{(\alpha-1)/\alpha}\Big)\) \\ without light-tailed assumptions on covariates?}
\end{center}
We answer this question affirmatively: under the standard finite fourth-moment covariate assumption, which is common even in finite-variance noise settings, we give matching upper and lower bounds on the estimation error, provided that the sample size is sufficiently large, without knowing $\alpha$ or $\nu$.

Our main algorithm is a black-box reduction from weak-moment regression to
finite-variance regression.
Starting from a sufficiently accurate 
initialization, our method prunes the most extreme sample points using an estimated threshold, and passes the resulting sample set
to a black-box finite-variance regression
subroutine.
The role of pruning is to make the sample distribution effectively have a finite variance.
The pruning threshold is chosen to balance between the variance reduction and the introduced bias, while our particular empirically-driven threshold-estimation procedure ensures that the threshold can be computed without knowledge of $\alpha$ and $\nu$.
On top of this pruning procedure, we also employ a standard iterative approach for regression: at every stage with a current regression vector estimate $\widehat{\theta}$, we compute the residual $\tilde{y} = y - \widehat{\theta}^\top x $ and run regression on the residual to further reduce the error in $\widehat{\theta}$.

Our reduction is computationally efficient, guaranteeing the polynomial runtime of the overall regression algorithm whenever the black-box finite-variance regression subroutine has polynomial runtime. The main guarantees are summarized  in the following proposition; here and subsequently, we write $\Sigma:=\E[xx^\top]-\E[x]\E[x]^\top$ and $\|v\|_\Sigma:=(v^\top\Sigma v)^{1/2}$.

\begin{proposition}[Informal]
Under covariance $\Sigma$, and finite fourth-moment covariates, our black-box reduction
turns any finite-variance regression routine with sub-Gaussian rate
into an estimator $\widehat\theta$ satisfying, above an explicit sample-size
threshold,
\begin{equation}
\label{eq:intro-estimation-rate}
    \|\widehat\theta-\theta^*\|_\Sigma
    \lesssim_\alpha
    \nu^{1/\alpha}
    \left(\frac{d+\log(1/\delta)}{n}\right)^{\frac{\alpha-1}{\alpha}}
\end{equation}
with probability at least $1-\delta$. A matching lower bound under the same conditions shows that
this rate is minimax optimal up to constants.
\end{proposition}

Without covariate assumptions, parameter estimation is not identifiable in
general. However, we can still obtain a parallel black-box reduction for prediction if the \emph{non-central} weak-moment of the response is bounded by $\nu$.

\begin{proposition}[Informal]
For arbitrary covariates, under the stronger assumption $ \E[|y|^\alpha \mid x] \le \nu$, our reduction combined with a distribution-free finite-variance prediction subroutine returns
a predictor $\widehat h$ satisfying, for sample size threshold $n \ge 2d$,
\begin{equation}
\label{eq:intro-df-rate}
    \left(\E_x(\widehat h(x)-x^\top\theta^*)^2\right)^{1/2}
    \lesssim_\alpha
    \nu^{1/\alpha}
    \left(
    \frac{d\log(n/d)+\log(1/\delta)}{n}
    \right)^{\frac{\alpha-1}{\alpha}},
\end{equation}
where the additional $\log(n/d)$ factor is inherited from the finite-variance distribution-free subroutine.
\end{proposition}

\paragraph{Contributions.}
We summarize our main contributions as follows.
\begin{itemize}[leftmargin=5ex]
    \item We introduce a black-box reduction for solving regression with only a weak
    error moment via finite-variance regression.  The reduction treats the
    finite-variance estimator as a subroutine and does not require knowledge of the
    moment exponent $\alpha$, scale $\nu$, or covariance $\Sigma$.

    \item Under a finite fourth-moment condition on the design, any
    subroutine for the finite-variance setting attaining a sub-Gaussian rate yields the rate in 
    \pref{eq:intro-estimation-rate} when $n$ is above a suitably-defined threshold.
    The resulting estimator is computationally efficient whenever the
    finite-variance subroutine is computationally efficient.

    \item We prove matching lower bounds under the same assumptions.  In our construction, the covariates have covariance $\Sigma$ and satisfy the fourth-to-second moment comparison used by the upper bound, showing that the rate \pref{eq:intro-estimation-rate} is
    minimax optimal up to 
    constants (at least for $n$ exceeding the above-mentioned threshold).

    \item For distribution-free prediction with arbitrary covariates, when the weak moment condition is imposed on the full response rather than only on the regression error, our reduction gives the infinite-variance
  analogue of the best known finite-variance guarantee in \pref{eq:intro-df-rate}; the additional $\log(n/d)$ factor is inherited from that finite-variance benchmark.
\end{itemize}
We also provide a simple experiment in \pref{app:numerical-experiment} regarding known vs.~unknown parameters $(\alpha,\nu)$.

\begin{table}[t]
\centering
\footnotesize
\caption{Representative high-probability robust regression guarantees.}
\definecolor{TableGreen}{RGB}{22,125,70}
\definecolor{TableRed}{RGB}{180,55,45}
\definecolor{TableGold}{RGB}{170,115,0}
\newcommand{\tabyes}{\textcolor{TableGreen}{\cmark}}
\newcommand{\tabno}{\textcolor{TableRed}{\xmark}}
\newcommand{\tabmaybe}{\textcolor{TableGold}{\textbf{?}}}
\newcommand{\tabdash}{\textcolor{black!55}{-}}
\setlength{\tabcolsep}{2.3pt}
\renewcommand{\arraystretch}{1.08}
\begin{threeparttable}
\begin{tabular}{@{}p{0.267\textwidth}
                >{\centering\arraybackslash}p{0.177\textwidth}
                >{\centering\arraybackslash}p{0.18\textwidth}
                >{\centering\arraybackslash}p{0.085\textwidth}
                >{\centering\arraybackslash}p{0.075\textwidth}
                >{\centering\arraybackslash}p{0.106\textwidth}@{}}
\toprule
Work & \makecell{Heavy-tailed\\covariates} & \makecell{Infinite variance\\$(\alpha\in(1,2))$ moment}
& Adaptive & Efficient & \makecell{\textbf{Statistical}\\\textbf{optimality}} \\
\midrule
\citet{sun2020adaptive}
& \tabno
& \tabyes
& \tabno & \tabyes & \tabyes \\
\citet{el2024minimax}
& 4th moment
& \tabno
& \tabdash & \tabno & \tabyes \\
\citet{lugosi2019risk,oliveira2025trimmed}
& SB or $L_2/L_1$
& \tabno
& \tabdash & \tabno & \tabyes \\
\cite{pmlr-v32-hsu14,depersin2020spectral,cherapanamjeri2020algorithms}
& $L_4/L_2$ or stronger
& \tabno
& \tabdash & \tabyes & \tabyes \\
\citet{pensia2025robust}
& $L_4/L_2$
& independent noise\tnote{b}
& \tabdash & \tabyes & \tabyes \\
\citet{mourtada2022distribution}
& arbitrary
& raw 2nd
& \tabdash & \tabno & \tabmaybe\tnote{a} \\
\citet{sun2020adaptive}
& bounded + Gram
& \tabyes
& \tabno & \tabyes & \tabno \\
Coordinatewise
& covariance
& \tabyes
& \tabyes & \tabyes & \tabno \\
\midrule
\multirow{3}{*}{This paper}
& 4th moment
& \tabyes
& \tabyes & \tabno & \tabyes\tnote{c} \\
& $L_4/L_2$ or stronger
& \tabyes
& \tabyes & \tabyes\tnote{c} & \tabyes\tnote{c} \\
& arbitrary
& raw $\alpha$
& \tabyes & \tabno & \tabmaybe\tnote{a} \\
\bottomrule
\end{tabular}
\begin{tablenotes}[flushleft]
\footnotesize
\item[] ``Adaptive'' means adaptation to the weak-moment exponent and scale.
\item[a] Distribution-free prediction has an additional $\log(n/d)$ factor
relative to the sub-Gaussian prediction rate.
\item[b] Assumes i.i.d. errors independent of the covariates and calibrates the
guarantee through a noise quantile.
\item[c] Assuming an efficient, optimal black-box subroutine.
\end{tablenotes}
\end{threeparttable}
\label{tab:comparison}
\end{table}

\subsection{Related Work}

We outline the most relevant prior works here, and give a more detailed overview in \pref{app:further_related}. 

\paragraph{Mean estimation.}
Robust mean estimation provides a canonical setting for high-probability estimation under heavy tails; see \citet{lugosi2019} for a survey.  Median and
median-of-means ideas have a long history
\citep{mediansurvay,minsker2015geometric}, and Catoni's work gives
exponential-type deviation bounds under weak moment assumptions
\citep{Catoni2012}.  The finite-variance case is now well understood in one
dimension and high dimensions \citep{Bub13a, devroye2015subgaussianmeanestimators,lee2022optimal,
LugosiMendelson2019,pmlr-v99-cherapanamjeri19b,hopkins2020mean,lee2022optimalhigh}.  The
infinite-variance case has more recently been developed for both optimal rates
and adaptation to the unknown moment exponent or scale
\citep{cherapanamjeri2022optimal,lee2025all,dang2023optimality,
whitehouse2024mean}.  Related weak-moment ideas also appear in streaming
estimation and bandits \citep{Bub13a,das2024near,pmlr-v247-genalti24a}.  The
regression problem considered here adds a covariate distribution and requires
controlling prediction or parameter error rather than only a mean.

\paragraph{Heavy-tailed Regression.}
Least squares can be highly suboptimal under heavy tails
\citep{Brownlees_2015,lugosi2019}.  Median-of-means-based methods provide one
route to high-probability regression under weak moments
\citep{pmlr-v32-hsu14,hsu2016loss,lugosi2019risk,lecue2020robust}; however, the
statistically optimal tournament-style guarantees are not computationally
efficient, while computationally tractable variants typically require stronger
structure or lose in the rate and/or confidence dependence.

Truncated least-squares and Catoni-style robust-risk objectives control
heavy-tailed losses by replacing empirical means in the objective
\citep{audibert2011robust,Brownlees_2015}; trimmed-mean procedures give
uniform guarantees for robust mean estimation and quadratic regression under
weak moment and small-ball assumptions \citep{oliveira2025trimmed}.  The
finite-variance quantile-risk benchmark under fourth moments is due to
\citet{el2024minimax}.  On the computational side, Huber-type methods give
sharp dense rates under sub-Gaussian designs and extend to heavy-tailed
covariates through covariate filtering \citep{sun2020adaptive,pensia2025robust}.
Spectral and semidefinite methods give computationally efficient
high-probability guarantees under $L_4/L_2$, certifiable hypercontractivity,
bounded covariance, or related assumptions
\citep{depersin2020spectral,oliveira2022spectral,
cherapanamjeri2020algorithms}.  Thus, the finite-variance literature separates
statistical optimality under weak assumptions from computational efficiency
under stronger covariate regularity.

\section{Black-Box Reduction Algorithms}

\subsection{Estimation Under Finite Fourth-Moment Design}
This section gives a black-box reduction from finite-variance regression
to regression with only a conditional $\alpha$-moment of the noise.

\begin{assumption}[Fourth-to-second moment comparison]
\label{assum:fourth-moment-design}
The covariance $\Sigma=\E[xx^\top]$ is positive definite, and
\[
    K_4:=
    \sup_{\|v\|_\Sigma=1}\E[(v^\top x)^4]<\infty.
\]
\end{assumption}

We will see that a ``pruning-bias'' step in our analysis requires control of
$\E|v^\top x|^{2\alpha}$ for $\|v\|_\Sigma=1$, and 
\pref{assum:fourth-moment-design} controls these quantities uniformly over $\alpha\in(1,2]$, giving a common covariate assumption for adaptation to unknown $\alpha$. 
Note that $K_4$ may depend on $d$, unlike the constant hypercontractivity assumption used in prior work \citep{cherapanamjeri2020algorithms, depersin2020spectral, pensia2025robust}.
In particular, for covariates sampled uniformly from the signed standard basis vectors,
this constant grows linearly with dimension.

\begin{assumption}[Finite-variance estimation primitive]
\label{assum:finite-variance-estimation}
There is a regression routine $\mathsf{EstFV}$ with a rate function
$R_{\rm est}$ satisfying the following:  Whenever
\[
    z=x^\top g+\xi,
    \qquad
    \E[x\xi]=0,
    \qquad
    \E[\xi^2]\le\sigma^2,
\]
the output of $\mathsf{EstFV}$ on $N$ i.i.d. samples with failure probability
$\eta$ satisfies
\[
    \|\widehat g-g\|_\Sigma
    \le
    \sigma R_{\rm est}(N,\eta)
\]
with probability at least $1-\eta$, for some rate function $R_{\rm est}$ that is nonincreasing in both
arguments.
\end{assumption}

\begin{algorithm}[t]
\caption{Black-Box Weak-Moment Regression}
\label{alg:iterated-pruning}
\DontPrintSemicolon
\SetKwInput{KwIn}{Input}
\SetKwInput{KwOut}{Output}

\KwIn{Samples $\{(x_i,y_i)\}_{i=1}^n$, confidence $\delta\in(0,1/10)$, finite-variance routine $\mathsf{EstFV}$, and rate function $R_{\rm est}$}
\KwOut{Estimator $\widehat\theta$}

$S_0,\{(T_j,P_j)\}_{j=1}^{\log n+1}
\leftarrow
\mathsf{Split}(\{(x_i,y_i)\}_{i=1}^n)$ \tcp*{Satisfying \pref{eq:split-size}}

$\delta_j\leftarrow\delta/(8\log n)\ (j\le\log n)$,
$\delta_{\log n+1}\leftarrow\delta/8$\;

$\widehat\theta_0\leftarrow \mathsf{InitLAD}(S_0,\delta)$\ (see \pref{alg:polynomial-warm-start});

\For{$j=1,\ldots,\log n+1$}{
    $\widehat\tau_j\leftarrow
    \inf\left\{t:
    \frac1{|T_j|}
    \sum_{(x_i,y_i)\in T_j}
    \mathbf 1\{|y_i-x_i^\top\widehat\theta_{j-1}|\le t\}
    \ge
    1-64\left(
    R_{\rm est}(|P_j|,\delta_j)^2+
    \frac{\log(1/\delta_j)}{|T_j|}
    \right)
    \right\}$\;

    \ForEach{$(x_i,y_i)\in P_j$}{
        $z_i\leftarrow (y_i-x_i^\top\widehat\theta_{j-1})
        \mathbf 1\{|y_i-x_i^\top\widehat\theta_{j-1}|\le\widehat\tau_j\}$\;
    }

    $\widehat g_j\leftarrow
    \mathsf{EstFV}\left(\{(x_i,z_i):(x_i,y_i)\in P_j\},
    \delta_j\right)$\;

    $\widehat\theta_j\leftarrow\widehat\theta_{j-1}+\widehat g_j$\;
}

\Return{$\widehat\theta_{\log n+1}$}\;
\end{algorithm}

Our procedure, described in \pref{alg:iterated-pruning},  does not require knowing $\alpha$, $\nu$, $\Sigma$, or $K_4$.  The unknown moment exponent and scale enter only
through the analysis; the empirical thresholds are estimated from the data.
The split in \pref{alg:iterated-pruning} is chosen so that
\begin{equation}
\label{eq:split-size}    
    |S_0|\ge \frac{n}{4},\qquad
    |T_j|,|P_j|\ge \frac{n}{8\log n}\quad (j\le\log n),
    \qquad
    |T_{\log n+1}|,|P_{\log n+1}|\ge \frac{n}{4}.
\end{equation}
Throughout the paper, we refer to $T_j$ and $P_j$ as \emph{threshold blocks} and \emph{pruning blocks} respectively.  The main algorithmic steps are as follows:
\begin{itemize}[leftmargin=*]
    \item \textbf{Warm-start initialization.}
    The black-box reduction only needs an initial estimate at constant scale.
    We use \pref{alg:polynomial-warm-start}, which computes Least Absolute Deviations (LAD) estimates on
    independent subblocks and returns the candidate with the smallest median
    validation LAD loss.

    \item \textbf{Threshold estimation.}
    At each iteration, we use an independent threshold block to choose a threshold for the absolute residuals under the current estimate.
    The fraction of residuals to remove is chosen using the error bound for the finite-variance estimator. 
    This balances the bias from pruning with the estimation error on the retained observations, without knowing
  $\alpha$ or $\nu$.

    \item \textbf{Residual pruning.}
    On the corresponding pruning block, residuals above the estimated threshold are set to zero.
    Conditional on earlier rounds and the current threshold block, the
    transformed observations in the independent pruning block remain i.i.d.,
    as required by \pref{assum:finite-variance-estimation}, whereas reusing the same observations would require a guarantee
    that also accounts for the resulting dependence.
    The pruned residuals define a
    finite-variance regression problem whose regression parameter is close to
    the remaining error $\theta^*-\widehat\theta$.

    \item \textbf{Finite-variance correction.}
    The routine $\mathsf{EstFV}$ estimates this correction term $\theta^*-\widehat\theta$, and the estimate is
    updated.  The early iterations reduce the warm-start error to a constant
    scale; the final iteration gives the target
    $\nu^{1/\alpha}R_{\rm est}^{2(\alpha-1)/\alpha}$ rate.
\end{itemize}

\begin{algorithm}[t]
\caption{Initial Estimate via Least Absolute Deviations (InitLAD)}
\label{alg:polynomial-warm-start}
\DontPrintSemicolon
\SetKwInput{KwIn}{Input}
\SetKwInput{KwOut}{Output}

\KwIn{Initialization block $S_0$ and confidence $\delta\in(0,1/10)$}
\KwOut{Initial estimate $\widehat\theta_0$}

$M\leftarrow\left\lceil512\log\frac1\delta\right\rceil$\;

$\{B_k\}_{k=1}^M,\{V_g\}_{g=1}^M\leftarrow\mathsf{Split}(S_0)$ \tcp{Equal size blocks}

\For{$k=1,\ldots,M$}{
    $\widehat\theta_k\leftarrow
    \displaystyle\argmin_{\beta\in\bbR^d}
    \sum_{(x_i,y_i)\in B_k} |y_i-x_i^\top\beta|$\;

    $\widehat L_k\leftarrow
    \displaystyle\operatorname{median}_{g\in[M]}
    \frac1{|V_g|}
    \sum_{(x_i,y_i)\in V_g}|y_i-x_i^\top\widehat\theta_k|$\;
}

\Return{$\widehat\theta_{k^*}$ where $k^* = \argmin_{k\in[M]}\widehat L_k$}\;
\end{algorithm}

\begin{theorem}[Generic estimation theorem]
\label{thm:generic-estimation}
Under \pref{assum:error-moment}, \pref{assum:fourth-moment-design}, and
\pref{assum:finite-variance-estimation}, there exist constants
$A_0,B_0>0$ such that if
\begin{equation}
\label{eq:generic-warm-start-sample}
    n
    \ge
    A_0K_4^2d\log(1/\delta),
\end{equation}
and
\begin{equation}
\label{eq:generic-contraction-condition}
    R_{\rm est}\left(\frac{n}{8\log n},\frac{\delta}{8\log n}\right)^2
    +
    \frac{\log n\,\log((\log n)/\delta)}{n}
    \le
    \left(\frac{1}{A_0\sqrt{K_4}}\right)^{\frac{\alpha}{\alpha-1}},
\end{equation}
then \pref{alg:iterated-pruning} satisfies, with probability at least
$1-\delta$,
\begin{equation}
\label{eq:generic-estimation-bound}
    \|\widehat\theta-\theta^*\|_\Sigma
    \le
    B_0\nu^{1/\alpha}
    \left(
    R_{\rm est}\left(\frac{n}{4},\frac{\delta}{8}\right)^2
    +
    \frac{\log(1/\delta)}{n}
    \right)^{\frac{\alpha-1}{\alpha}}.
\end{equation}
\end{theorem}

\begin{corollary}[Fourth-moment plug-in]
\label{cor:hanchi-plug-in}
Assume $\mathsf{EstFV}$ is the fourth-moment finite-variance estimator of
\citet{el2024minimax}.  Let $N_{\rm EH}(d,\delta,\calD_X)$ denote the
sample-size lower bound in their theorem at confidence $\delta$ for design distribution $\calD_X$,
and let $C_{\rm est}$ be such that
\[
    R_{\rm est}(N,\eta)
    \le
    C_{\rm est}\sqrt{\frac{d+\log(1/\eta)}{N}}.
\]
There exist constants $A_{\alpha,C_{\rm est}},B_{\alpha,C_{\rm est}}>0$,
depending only on $\alpha$ and $C_{\rm est}$, such that if\footnote{The notation $a \vee b$ means $\max\{a,b\}$.}
\[
    \begin{aligned}
    n
    &\ge
    A_{\alpha,C_{\rm est}}
    \Big[
        K_4^{\,3\vee\left(1+\frac{\alpha}{2(\alpha-1)}\right)}
        \left(d+\log(1/\delta)\right)
        \log\left(\frac{d+\log(1/\delta)}{\delta}\right)
        \\
        &\quad+
        N_{\rm EH}(d,\delta,\calD_X)
        \log\left(\frac{N_{\rm EH}(d,\delta,\calD_X)}{\delta}\right)
    \Big],
    \end{aligned}
\]
then \pref{alg:iterated-pruning} satisfies, with probability at least $1-\delta$,
\begin{equation}
\label{eq:hanchi-plugin-bound}
    \|\widehat\theta-\theta^*\|_\Sigma
    \le
    B_{\alpha,C_{\rm est}}\nu^{1/\alpha}
    \left(\frac{d+\log(1/\delta)}{n}\right)^{\frac{\alpha-1}{\alpha}}.
\end{equation}
\end{corollary}

\begin{remark}[Computational efficiency]
\pref{alg:polynomial-warm-start} is computationally efficient since it solves only
convex LAD optimizations.  The pruning rounds evaluate empirical quantiles at the
current estimate and call
$\mathsf{EstFV}$.  Hence \pref{alg:iterated-pruning} is computationally
efficient whenever the finite-variance subroutine is computationally efficient.
Using the finite-variance regression algorithm of
\citet{cherapanamjeri2020algorithms} as $\mathsf{EstFV}$ therefore gives a
computationally efficient version under their stronger certifiable hypercontractivity
assumptions, whereas using the regression algorithm of \cite{el2024minimax} as finite-variance subroutine gives a computationally inefficient algorithm.
\end{remark}

\begin{remark}[Empirical tuning]
Our algorithm adapts to the unknown moment exponent and scale through the
threshold blocks.  If $\alpha$ and $\nu$ are known, one may replace
these empirical thresholds by the deterministic thresholds used in the proof.
The same tuning idea can be used for weak-moment mean estimation: the
deterministic pruning radius in Algorithm 5 of
\citet{cherapanamjeri2022optimal} can be replaced by an empirical radius from
an independent block.  The four properties used in their Lemma C.7 continue to
hold, up to an additional threshold-estimation event; consequently, the
condition $\delta\ge e^{-\mathcal{O}(n)}$ only changes by numerical constants. See \pref{app:empirical-mean-pruning} for more details.
\end{remark}

\subsection{Prediction Under Arbitrary Covariates}

The distribution-free prediction version of our algorithm uses the same pruning step with the
zero initial predictor.  In this setting, we adopt a weak moment assumption on the response itself, rather than on the regression noise, as formalized in the following.

\begin{assumption}[Raw response moment]
\label{assum:raw-response-moment}
There is a parameter $\theta^*\in\bbR^d$ such that
\[
    \E[y\mid x]=x^\top\theta^*,
    \qquad
    \E\left[|y|^\alpha\mid x\right]\le\nu
\]
for some $1<\alpha\le2$ and $\nu>0$.
\end{assumption}

This is stronger than \pref{assum:error-moment} in the sense that a bound on the raw moment implies an analog on the centered moment (with at most a constant factor increase), but not vice versa.  
The reason for this stronger assumption is that in the distribution-free setting, a weak-moment assumption on the noise is not enough, since there exist noise-free instances such that the prediction error of every estimator is lower bounded by an absolute constant with constant probability (see Proposition 2 in \cite{mourtada2022distribution}).   We will see that the raw response moment directly controls the bias from pruning $y$, so  no warm start or iterative correction is needed.  This is in contrast with \pref{alg:iterated-pruning},
where only the noise moment is bounded, and successive residual corrections progressively reduce
the contribution of $x^\top(\theta^*-\widehat\theta)$ to the pruning bias.

\begin{assumption}[Finite-variance prediction primitive]
\label{assum:finite-variance-prediction}
There is a prediction routine $\mathsf{PredFV}$ with rate function $R_{\rm pred}$ such that the following holds.
For any distribution of $(x,z)$ with $\E[z^2\mid x]\le\sigma^2$ a.s., the
output $\widetilde h$ of $\mathsf{PredFV}$ on $N$ i.i.d. samples with failure
probability $\eta$ satisfies
\[
    \E\big(\widetilde h(x) - z\big)^2
    -
    \inf_{\theta\in\bbR^d}
    \E\left(x^\top\theta - z\right)^2
    \le
    \sigma^2 R_{\rm pred}(N,\eta)^2
\]
with probability at least $1-\eta$, for some rate function $R_{\rm pred}$ that is nonincreasing in both
arguments.
\end{assumption}

\begin{algorithm}[t]
\caption{Black-Box Distribution-Free Weak-Moment Regression}
\label{alg:df-blackbox-prediction}
\DontPrintSemicolon
\SetKwInput{KwIn}{Input}
\SetKwInput{KwOut}{Output}

\KwIn{$2n$ samples $\{(x_i,y_i)\}_{i=1}^{2n}$, confidence $\delta\in(0,1)$, class $\calF$, finite-variance prediction routine $\mathsf{PredFV}$, and rate function $R_{\rm pred}$}
\KwOut{Predictor $\widehat h$}

$T,P\leftarrow\mathsf{Split}(\{(x_i,y_i)\}_{i=1}^{2n})$,
$|T|=|P|=n$\;

\If{$R_{\rm pred}(n,\delta/2)^2+\frac{\log(4/\delta)}{n}>1/128$}{
    \Return{$\widehat h\equiv0$}\;
}

$\widehat\tau\leftarrow
\inf\left\{t:
\frac1{|T|}
\sum_{(x_i,y_i)\in T}\mathbf 1\{|y_i|\le t\}
\ge
1-64\left(
R_{\rm pred}(n,\delta/2)^2+
\frac{\log(4/\delta)}{n}
\right)
\right\}$\;

\ForEach{$(x_i,y_i)\in P$}{
    $z_i\leftarrow y_i\mathbf 1\{|y_i|\le\widehat\tau\}$\;
}

$\widetilde h\leftarrow
\mathsf{PredFV}\left(\{(x_i,z_i):(x_i,y_i)\in P\},\delta/2\right)$\;

\Return{$x\mapsto
\max\{-\widehat\tau,\min\{\widehat\tau,\widetilde h(x)\}\}$}\;
\end{algorithm}

\begin{theorem}[Adaptive distribution-free prediction]
\label{thm:df-blackbox-prediction}
Suppose that \pref{assum:raw-response-moment} and
\pref{assum:finite-variance-prediction} hold, with no condition imposed on the
distribution of $x$. Then
\pref{alg:df-blackbox-prediction} satisfies with probability at least
$1-\delta$ that
\begin{equation}
\label{eq:df-prediction-bound}
    \left(\E_x(\widehat h(x)-x^\top\theta^*)^2\right)^{1/2}
    \le
    \sqrt{514}\,\nu^{1/\alpha}
    \left(
        R_{\rm pred}(n,\delta/2)^2+\frac{\log(4/\delta)}{n}
    \right)^{\frac{\alpha-1}{\alpha}}.
\end{equation}
\end{theorem}
If the squared risks $\E_{x,y}(\widehat h(x)-y)^2$ and
$\E_{x,y}(x^\top\theta^*-y)^2$ are finite, then \pref{eq:df-prediction-bound} also
implies the same bound on the squared excess risk, since 
$    \E_{x,y}(\widehat h(x)-y)^2
    -
    \E_{x,y}(x^\top\theta^*-y)^2
    =
    \E_x\big(\widehat h(x)-x^\top \theta^*\big)^2
$.

\begin{corollary}[Distribution-free weak-moment prediction]
\label{cor:df-blackbox-prediction}
Suppose that \pref{assum:raw-response-moment} holds, and let $\mathsf{PredFV}$ be the
distribution-free finite-variance predictor of
\citet{mourtada2022distribution}.  For $n\ge2d$, their theorem gives
\pref{assum:finite-variance-prediction} with 
\[
    R_{\rm pred}(n,\eta)^2
    \le
    C_{\rm df}
    \frac{d\log(n/d)+\log(1/\eta)}{n}
\]
for some numerical constant $C_{\rm df}$.  
Then \pref{alg:df-blackbox-prediction} satisfies, with probability at least
$1-\delta$,
\begin{equation}
\label{eq:df-blackbox-corollary-bound}
    \left(\E_x(\widehat h(x)-x^\top\theta^*)^2\right)^{1/2}
    \le
    \sqrt{514}\,(C_{\rm df}+1)^{\frac{\alpha-1}{\alpha}}
    \nu^{1/\alpha}
    \left(
        \frac{d\log(n/d)+\log(4/\delta)}{n}
    \right)^{\frac{\alpha-1}{\alpha}}.
\end{equation}
\end{corollary}

\subsection{Proofs}

We first state two key lemmas, whose proofs are deferred to Appendices \ref{app:proof_warm_start} and \ref{app:proof_residual_pruning}.

\begin{lemma}[Warm-start initialization]
\label{lem:warm-start-initialization}
Under \pref{assum:error-moment} and \pref{assum:fourth-moment-design}, if
\begin{equation}
\label{eq:warm-start-sample-size}
    |S_0|
    \ge
    2^{31} K_4^2d\log(1/\delta),
\end{equation}
then the output of \pref{alg:polynomial-warm-start} satisfies, with probability at least
$1-\delta/8$,
\begin{equation}
\label{eq:warm-start-bound}
    \|\widehat\theta_0-\theta^*\|_\Sigma
    \le
    2^{24}K_4^2\nu^{1/\alpha}.
\end{equation}
\end{lemma}

\begin{lemma}[Residual-pruning step]
\label{lem:one-pruning-step}
Suppose that \pref{assum:error-moment} and \pref{assum:fourth-moment-design} hold.  Let
$\bar\theta\in\bbR^d$ be deterministic, and let $T$ and $P$ be independent
blocks of size $N$.
Let $\widehat\tau$ be the empirical quantile of
$\{|y_i-x_i^\top\bar\theta|:(x_i,y_i)\in T\}$ at level
\begin{equation}
\label{eq:one-step-quantile-level}
    1-64\left(
        R_{\rm est}(N,\delta)^2+\frac{\log(1/\delta)}{N}
    \right).
\end{equation}
For $(x_i,y_i)\in P$, define
\[
    z_i:=(y_i-x_i^\top\bar\theta)
    \mathbf 1\{|y_i-x_i^\top\bar\theta|\le\widehat\tau\},
    \qquad
    \widehat g
    :=
    \mathsf{EstFV}\left(\{(x_i,z_i):(x_i,y_i)\in P\},\delta\right).
\]
If
\begin{equation}
\label{eq:one-step-small-gamma}
    R_{\rm est}(N,\delta)^2+\frac{\log(1/\delta)}{N}
    \le
    \frac{1}{128},
\end{equation}
and  $\mathsf{EstFV}$ satisfies \pref{assum:finite-variance-estimation} at block size $N$, then with probability at least
$1-2\delta$,
\begin{equation}
\label{eq:one-step-bound}
    \|\bar\theta+\widehat g-\theta^*\|_\Sigma
    \le
    32
    \left(
        1+
        \sqrt{K_4}\,
        \frac{\|\bar\theta-\theta^*\|_\Sigma}{\nu^{1/\alpha}}
    \right)
    \nu^{1/\alpha}
    \left(
        R_{\rm est}(N,\delta)^2+\frac{\log(1/\delta)}{N}
    \right)^{\frac{\alpha-1}{\alpha}}.
\end{equation}
\end{lemma}

\begin{proof}[Proof of \pref{thm:generic-estimation}]
We prove the theorem with $A_0=2^{40}$ and $B_0=2^8$.
Define the normalized radii
\[
    r_j=\frac{\|\widehat\theta_j-\theta^*\|_\Sigma}{\nu^{1/\alpha}}.
\]
By \pref{lem:warm-start-initialization} and
\pref{eq:generic-warm-start-sample}, with probability at least $1-\delta/8$, the choice of $\delta_j$ and the sizes in \pref{eq:split-size} give
\begin{equation}
\label{eq:initial-radius}
    r_0\le 2^{24} K_4^2.
\end{equation}
For each iterative round, \pref{lem:one-pruning-step} is applied with block
size at least $\frac{n}{8\log n}$ and confidence $\delta_j = \delta/(8\log n)$.
For every such round, we have
\begin{align}
    R_{\rm est}(|P_j|,\delta_j)^2+
    \frac{\log(1/\delta_j)}{|T_j|}
    &\le
    32\left[
    R_{\rm est}\left(\frac{n}{8\log n},\frac{\delta}{8\log n}\right)^2
    +
    \frac{\log n\,\log((\log n)/\delta)}{n}
    \right]
    \nonumber\\
    &\le
    32\left(\frac{1}{A_0\sqrt{K_4}}\right)^{\frac{\alpha}{\alpha-1}}
    \qquad \text{(by \pref{eq:generic-contraction-condition})}
    \nonumber\\
    &\le
    \frac{32}{A_0^2}\le\frac{1}{128}.
    \qquad\qquad\quad~ \text{($K_4\ge1$, $\alpha\le2$)}
    \label{eq:iterative-round-domination}
\end{align}
Therefore,
\begin{align}
    32\sqrt{K_4}
    \left(
    R_{\rm est}(|P_j|,\delta_j)^2+
    \frac{\log(1/\delta_j)}{|T_j|}
    \right)^{\frac{\alpha-1}{\alpha}}
    &\le
    \frac{32^{1+\frac{\alpha-1}{\alpha}}}{A_0}
    \nonumber\\
    &\le
    \frac{32\sqrt{32}}{A_0}\le\frac18.
    && \text{($\alpha\le2$)}
    \label{eq:iterative-pruning-size}
\end{align}
Then, by \pref{lem:one-pruning-step} and a union bound over the
iterative rounds, we have with probability at least $1-\delta/4$ that
\[
    r_{j+1}
    \le
    32\left(1+\sqrt{K_4}r_j\right)
    \left(
    R_{\rm est}(|P_j|,\delta_j)^2+
    \frac{\log(1/\delta_j)}{|T_j|}
    \right)^{\frac{\alpha-1}{\alpha}}
\]
in each iterative round.  Combining with \pref{eq:iterative-pruning-size}, we obtain
\begin{equation}
\label{eq:radius-recursion}
    r_{j+1}
    \le
    \frac{1}{8\sqrt{K_4}}+\frac{r_j}{8}.
\end{equation}
Let $r_{\mathrm{iter}}$ denote the normalized radius after the $\log n$ iterative rounds.  By repeated application of \pref{eq:radius-recursion} over these rounds
(along with $\sum_{i=1}^{\infty} 8^{-i} = 1/7$), we have
\begin{align}
\label{eq:iterated-radius}
    r_{\rm iter}
    \le
    \frac{1}{7\sqrt{K_4}}+8^{-\log_2 n}r_0
    \le
    \frac{2}{7\sqrt{K_4}}. && \text{(\pref{eq:initial-radius} and \pref{eq:generic-warm-start-sample})}
\end{align}

\pref{lem:one-pruning-step} is applied once more to the final threshold/pruning
pair with confidence $\delta/8$.  These blocks have a larger size of at least $n/4$, so the
monotonicity of $R_{\rm est}$ and \pref{eq:iterative-round-domination} imply that
the final-round quantity in \pref{eq:one-step-small-gamma} is at most
$1/128$.
The final-round concentration term is at most $8\log(1/\delta)/n$.  Using
\pref{eq:iterated-radius} and $B_0=2^8$, the final pruning step yields
\[
    \|\widehat\theta-\theta^*\|_\Sigma
    \le
    B_0\nu^{1/\alpha}
    \left(
    R_{\rm est}\left(\frac{n}{4},\frac{\delta}{8}\right)^2
    +
    \frac{\log(1/\delta)}{n}
    \right)^{\frac{\alpha-1}{\alpha}}
\]
with probability at least $1-\delta$.
\end{proof}

The proof of \pref{thm:df-blackbox-prediction} uses a similar empirical-threshold argument from
\pref{thm:generic-estimation}, but using the raw response rather than the residual.  After
pruning, $y\mathbf 1\{|y|\le\widehat\tau\}$ has bounded conditional second moment, so the finite-variance prediction primitive applies to its conditional
mean. The target shift caused by pruning is then controlled directly from the
conditional raw $\alpha$-moment. The full details, as well as the proofs of \pref{cor:hanchi-plug-in} and \pref{cor:df-blackbox-prediction}, are deferred to Appendices \ref{app:proof_prediction} and \ref{sec:proofs_corollaries}.

\section{Lower Bound}

Our main lower bound is stated as follows, and is proved in \pref{app:proof_lower}.

\begin{theorem}[Random-design lower bound]
\label{thm:random-design-lower}
Let $1<\alpha<2$, $\nu>0$, and let $\Sigma\in\bbR^{d\times d}$ be positive
definite.  If $n\ge d$, then there exists a finite parameter
set $\Theta$ and distributions $(\calD_\theta)_{\theta\in\Theta}$ satisfying
\pref{assum:error-moment}, and
\pref{assum:fourth-moment-design} with covariance $\Sigma$ and $K_4\le d$, 
such that
\[
    \inf_{\widehat{\theta}}\sup_{\theta\in\Theta}
    \Pr_{\calD_\theta^{\otimes n}}\left(
    \|\widehat{\theta}-\theta\|_\Sigma
    \ge
    \frac{1}{8}\nu^{1/\alpha}
    \left(\frac dn\right)^{\frac{\alpha-1}{\alpha}}
    \right)
    \ge
    \frac16,
\]
where the infimum is over all measurable estimators based on the
$n$ samples.

Moreover, the same family satisfies
\pref{assum:raw-response-moment}, and
\[
    \inf_{\widehat f}\sup_{\theta\in\Theta}
    \Pr_{\calD_\theta^{\otimes n}}\left(
    \left(\E_{x\sim\calD_X}
    (\widehat f(x)-x^\top\theta)^2\right)^{1/2}
    \ge
    \frac{1}{8}\nu^{1/\alpha}
    \left(\frac dn\right)^{\frac{\alpha-1}{\alpha}}
    \right)
    \ge
    \frac16
\]
where $\calD_X$ is the common marginal distribution of the covariates, and the infimum is over measurable predictors based on the $n$ samples.
\end{theorem}
\begin{sketch}
The lower bound uses a rare-response construction.  The parameter set becomes
a hypercube after applying the map $\theta\mapsto\Sigma^{1/2}\theta$, with
side length chosen so that $|x^\top\theta|$ is of order
$\nu^{1/\alpha}(d/n)^{(\alpha-1)/\alpha}$ on every design point.  The design
chooses one coordinate uniformly and with a random sign,
$x\in\{\pm\sqrt d\,\Sigma^{1/2}e_j:j\in[d]\}$.  This gives covariance
$\Sigma$ and satisfies the fourth-moment condition.  Conditional on $x$, the
response is usually zero and is equal to $(4n/d)x^\top\theta$ with probability
$d/(4n)$.  It follows that $\E[y\mid x]=x^\top\theta$, while the rare nonzero value is
large enough to make the conditional $\alpha$-moment of order $\nu$.

The statistical hardness comes from hiding each coordinate sign behind a rare
event.  If two parameters differ only in the $j$-th coordinate after applying
$\theta\mapsto\Sigma^{1/2}\theta$, their
one-sample laws differ only when the design selects coordinate $j$ and the
nonzero response occurs.  Hence, the total variation distance between the
$n$-sample laws is at most $1/4$.  Assouad's method then implies that, under the
uniform prior on the hypercube, any estimator must make a constant fraction of
coordinate sign mistakes on average.  Each wrong sign costs one side length in
the corresponding coordinate after applying $\theta\mapsto\Sigma^{1/2}\theta$.
Summing these errors gives the
$\Sigma$-norm lower bound at the stated rate.  The same sign mistakes give the
direct $L_2(\calD_X)$ prediction lower bound.

The complete proof is given in \pref{app:proof_lower}.
\end{sketch}

\begin{remark}[Confidence term]
\pref{thm:random-design-lower} proves a dimension-dependent lower bound but does not capture the dependence on the confidence parameter $\delta$.  To obtain a confidence-dependent term, we can make use of the one-dimensional weak-moment
mean-estimation lower bound.  Specifically, take $Z$ uniform on $\{\pm1\}^d$, set
$x=\Sigma^{1/2}Z$, and restrict the parameter set to
$\theta_t=t\Sigma^{-1/2}e_1$.  Then $y=tZ_1+\varepsilon$, and multiplying by the
observed sign $Z_1$ gives the one-dimensional mean-estimation problem
$Z_1y=t+Z_1\varepsilon$ with the same conditional $\alpha$-moment.  Therefore,
there exists a constant $c_\alpha>0$, depending only on $\alpha$, such that the
lower bound of \citet{devroye2015subgaussianmeanestimators} implies, for
$0<\delta<c_\alpha$,
\[
    \inf_{\widehat{\theta}}\sup_t
    \Pr\left(
        \|\widehat{\theta}-\theta_t\|_\Sigma
        \ge
        c_\alpha\nu^{1/\alpha}
        \left(\frac{\log(1/\delta)}{n}\right)^{\frac{\alpha-1}{\alpha}}
    \right)
    \ge
    \delta.
\]
Combining this one-dimensional reduction with \pref{thm:random-design-lower}
gives the overall lower-bound rate with $d+\log(1/\delta)$ replacing the above numerator.
\end{remark}

\begin{remark}
When $n<d$, the same lower bounds hold with
$\left(d/n\right)^{(\alpha-1)/\alpha}$ replaced by $1$.  This follows by applying
the theorem at sample size $d$ and using that minimax risk cannot increase when
additional samples are given.
\end{remark}

For the constructed lower bound instance, $K_4\le d$ as stated in \pref{thm:random-design-lower}.  Hence, for any sample size of at least $n = \widetilde{\Omega}(d^3)$, our upper
bound is tight up to constants on this hard instance.

\section{Conclusion}

We introduced a black-box reduction for using finite-variance regression
procedures in the infinite-variance setting without knowing the moment parameters $(\alpha,\nu)$. The reduction does not require knowledge of the covariance matrix,
and it preserves the computational efficiency of the chosen subroutine.
Under the fourth-moment covariate condition, an optimal finite-variance
subroutine yields the minimax estimation rate. An analogous reduction gives
adaptive prediction guarantees for arbitrary covariates under a moment
bound on the raw response. 
For estimation, it remains open whether the fourth-moment condition on the covariates can
be weakened while retaining the same estimation rate.  For distribution-free prediction, the additional
$\log(n/d)$ factor is inherited from the finite-variance subroutine, and it remains open whether it can be removed.

\section*{Acknowledgement}

This work is supported by the Singapore National Research Foundation (NRF) and the Ministry of Digital Development and Information (MDDI) under the AI Visiting Professorship programme (Award AIVP-2024-003).

\bibliography{minimax-regression/mmbiblio}
\bibliographystyle{apalike}

\appendix
\newpage

\section{Proof of \pref{lem:warm-start-initialization} (Warm-Start Initialization)} \label{app:proof_warm_start}
\begin{proof}
We refer to $B_k$ and $V_g$ in \pref{alg:polynomial-warm-start} as LAD blocks and validation blocks, respectively.

The Paley--Zygmund inequality applied to $(x^\top v)^2$ implies, for every $v\in\bbR^d$,
\begin{equation}
\label{eq:paley-small-ball}
    \Pr\left(|x^\top v|\ge \frac12\|v\|_\Sigma\right)
    \ge
    \frac{9}{16K_4}.
\end{equation}
Let $M=\lceil512\log(1/\delta)\rceil$.  The split in
\pref{alg:polynomial-warm-start} is chosen so that
$|B_k|,|V_g|\ge |S_0|/(4M)$; by \eqref{eq:warm-start-sample-size},
$|B_k|,|V_g|\ge2^{20}K_4^2d$ for all $k,g\in[M]$.

For a LAD block $B_k$, let $\calE_k$ be the event that
\begin{equation}
\label{eq:lad-block-event}
    \frac1{|B_k|}\sum_{(x_i,y_i)\in B_k}|y_i-x_i^\top\theta^*|
    \le 16\nu^{1/\alpha}
    \quad\text{and}\quad
    \inf_{v\ne0}
    \frac1{|B_k|}
    \sum_{(x_i,y_i)\in B_k}
    \mathbf 1\left\{|x_i^\top v|\ge\frac12\|v\|_\Sigma\right\}
    \ge
    \frac{9}{32K_4}.
\end{equation}
By \pref{assum:error-moment} and Markov's inequality, we have
\begin{align}
    \Pr\left(
    \frac1{|B_k|}\sum_{(x_i,y_i)\in B_k}|y_i-x_i^\top\theta^*|
    >16\nu^{1/\alpha}
    \right)
    &\le
    \frac{\E|\varepsilon|}{16\nu^{1/\alpha}}
    \le
    \frac1{16}.
\end{align}
For the empirical small-ball part of \eqref{eq:lad-block-event}, let
\[
    \calC=
    \left\{
    \left\{x:|x^\top v|\ge\frac12\|v\|_\Sigma\right\}:v\ne0
    \right\}.
\]
Each $C\in\calC$ is a union of two halfspaces.  Halfspaces in $\bbR^d$
have VC dimension $d+1$, so Sauer's lemma bounds the number of subsets
induced by $\calC$ on $2|B_k|$ points by
$(2e|B_k|/(d+1))^{2d+2}$.  Letting $p=9/(16K_4)$,
\eqref{eq:paley-small-ball} gives $\Pr(C)\ge p$ for every $C\in\calC$.
Using $|B_k|\ge2^{20}K_4^2d$, the VC inequality for relative deviation
\citep[Eq.~(1)]{bartlett1999inequality} gives
\begin{align}
    &\Pr\left(
    \exists C\in\calC:\;
    \frac1{|B_k|}\sum_{(x_i,y_i)\in B_k}\mathbf 1\{x_i\in C\}
    <
    \frac12\Pr(C)
    \right)
    \nonumber\\
    &\qquad\le
    8\left(\frac{6|B_k|}{d+1}\right)^{2d+2}
    \exp\left(-\frac{|B_k|p}{32}\right)
    \le
    \frac1{64}.
    \label{eq:small-ball-empirical}
\end{align}
Hence, $\Pr(\calE_k)\ge59/64$.

We claim that on $\calE_k$, every $\beta$ with
$\|\beta-\theta^*\|_\Sigma\ge2^{12}K_4\nu^{1/\alpha}$ has larger empirical
LAD loss than $\theta^*$.  Let $r=\|\beta-\theta^*\|_\Sigma$.  By
\eqref{eq:lad-block-event}, at least a fraction $9/(32K_4)$ of the block
satisfies $|x_i^\top(\beta-\theta^*)|\ge r/2$, while
\[
    \frac{1}{|B_k|}
    \sum_{(x_i,y_i)\in B_k}
    \mathbf 1\{|y_i-x_i^\top\theta^*|>r/4\}
    \le
    \frac{4}{r}
    \frac1{|B_k|}\sum_{(x_i,y_i)\in B_k}|y_i-x_i^\top\theta^*|
    \le
    \frac{64\nu^{1/\alpha}}{r}
    \le
    \frac{1}{64K_4}.
\]
Hence, at least a fraction $1/(4K_4)$ of the block satisfies both
$|x_i^\top(\beta-\theta^*)|\ge r/2$ and
$|y_i-x_i^\top\theta^*|\le r/4$.  On these samples,
$|y_i-x_i^\top\beta|\ge r/4$, so the empirical LAD loss of $\beta$ is at
least $r/(16K_4)$, whereas that of $\theta^*$ is at most
$16\nu^{1/\alpha}$.  Since $r \ge 2^{12}K_4\nu^{1/\alpha}$, the former loss is larger as claimed.

Since $\widehat\theta_k$ is a LAD minimizer, it follows that $\calE_k$ implies
\begin{equation}
\label{eq:lad-candidate-accuracy}
    \|\widehat\theta_k-\theta^*\|_\Sigma
    \le 2^{12}K_4\nu^{1/\alpha}.
\end{equation}

The events $\calE_1,\ldots,\calE_M$ are independent, and our choice of $M$ gives
\[
    \Pr\left(\bigcap_{k=1}^M\calE_k^c\right)
    \le
    \left(\frac{5}{64}\right)^M
    \le
    \frac{\delta}{16}.
\]
On $\bigcup_{k=1}^M\calE_k$, let $k_0$ be the first index satisfying
\eqref{eq:lad-candidate-accuracy}.  For every validation block $V_g$ and every
$\beta\in\bbR^d$, let
\[
    L_g(\beta)
    =
    \frac1{|V_g|}
    \sum_{(x_i,y_i)\in V_g}|y_i-x_i^\top\beta|,
    \qquad
    \widehat L(\beta)=\operatorname{median}_{g=1,\ldots,M}L_g(\beta).
\]
After the LAD-block observations are realized on $\bigcup_{k=1}^M\calE_k$,
the candidate $\widehat\theta_{k_0}$ is fixed and the validation blocks remain
independent.  For a fresh validation block,
\begin{align*}
    \E L_g(\widehat\theta_{k_0})
    \le
    \E|\varepsilon|
    +
    \E|x^\top(\widehat\theta_{k_0}-\theta^*)|
    \le
    2^{13}K_4\nu^{1/\alpha}. \tag{by \pref{eq:lad-candidate-accuracy}, $\E|\varepsilon|\le \nu^{1/\alpha}$, and $\E|x^\top v|\le \|v\|_\Sigma$}
\end{align*}
Then, over the validation blocks, 
\begin{align}
\label{eq:good-validation-score}
    \Pr\left(
    \widehat L(\widehat\theta_{k_0})
    >
    2^{18}K_4\nu^{1/\alpha}
    \right)
    &\le
    \exp(-M/128)
    \le
    \frac{\delta}{32}.
    && \text{(Markov on each } V_g\text{ then Hoeffding)}
\end{align}

For each validation block $V_g$, define $\calF_g$ by replacing $B_k$ by
$V_g$ in \eqref{eq:lad-block-event}.  The analysis of $\calE_k$ applies verbatim to $\calF_g$, yielding $\Pr(\calF_g)\ge59/64$, and hence
\begin{align}
\label{eq:many-validation-blocks}
    &\Pr\left(
    \sum_{g=1}^M\mathbf 1\{\calF_g\}<\frac34M
    \right) \le
    \exp(-M/128)
    \le
    \frac{\delta}{32}.
    && \text{(Hoeffding)}
\end{align}
The analysis of $\calE_k$ further establishes the following analog on $\calF_g$: Every $\beta$ with
$\|\beta-\theta^*\|_\Sigma\ge2^{24}K_4^2\nu^{1/\alpha}$ satisfies
$L_g(\beta)\ge2^{20}K_4\nu^{1/\alpha}$.  Similarly, for
$r=\|\beta-\theta^*\|_\Sigma$,
at least a fraction $9/(32K_4)$ of the validation block has
$|x_i^\top(\beta-\theta^*)|\ge r/2$, while
\[
    \frac1{|V_g|}
    \sum_{(x_i,y_i)\in V_g}
    \mathbf 1\{|y_i-x_i^\top\theta^*|>r/4\}
    \le
    \frac{64\nu^{1/\alpha}}{r}
    \le
    \frac{1}{64K_4}.
\]
Hence, at least a fraction $1/(4K_4)$ of the validation block satisfies
$|y_i-x_i^\top\beta|\ge r/4$, and so
$L_g(\beta)\ge r/(16K_4)\ge2^{20}K_4\nu^{1/\alpha}$.

On $\bigcup_{k=1}^M\calE_k$ and outside the failure events in \eqref{eq:good-validation-score} and
\eqref{eq:many-validation-blocks}, every
$\beta$ with
$\|\beta-\theta^*\|_\Sigma\ge2^{24}K_4^2\nu^{1/\alpha}$ has
$\widehat L(\beta)>2^{18}K_4\nu^{1/\alpha}$, while
$\widehat L(\widehat\theta_{k_0})\le2^{18}K_4\nu^{1/\alpha}$.  The minimizer in
\pref{alg:polynomial-warm-start} therefore satisfies
\eqref{eq:warm-start-bound}.  The failure probabilities in
\eqref{eq:good-validation-score}, \eqref{eq:many-validation-blocks}, and
$\bigcap_{k=1}^M\calE_k^c$ sum to at most $\delta/8$.
\end{proof}

\section{Proof of \pref{lem:one-pruning-step} (Residual-Pruning Step)} \label{app:proof_residual_pruning}
\begin{proof}
Recall that we refer to $T_j$ and $P_j$ in \pref{alg:iterated-pruning} as \emph{threshold blocks} and \emph{pruning blocks} respectively.  
Let
\begin{equation}
\label{eq:one-step-gamma}
    \gamma=
        R_{\rm est}(N,\delta)^2+\frac{\log(1/\delta)}{N},
    \qquad
    \Delta=\bar\theta-\theta^*.
\end{equation}
By \eqref{eq:one-step-small-gamma}, $\gamma\le1/128$.  For a fresh sample $(x,y)$, the definition of $\Delta$ gives
$y-x^\top\bar\theta=\varepsilon-x^\top\Delta$.  Using $1<\alpha\le 2$ and Jensen's inequality, we have
\begin{align}
    \E|\varepsilon-x^\top\Delta|^\alpha
    &\le
    2\left(\E|\varepsilon|^\alpha+
    \E|x^\top\Delta|^\alpha\right) \nonumber\\
    &\le
    2\left(\nu+\|\Delta\|_\Sigma^\alpha\right).
    \label{eq:residual-alpha-moment}
\end{align}
For every $v$ with $\|v\|_\Sigma=1$, we have
\begin{align}
    &\E\left[|v^\top x|^\alpha|\varepsilon-x^\top\Delta|^\alpha\right] \nonumber \\
    &\le
    2\E\left[|v^\top x|^\alpha|\varepsilon|^\alpha\right]
    +2\E\left[|v^\top x|^\alpha|x^\top\Delta|^\alpha\right]
    \nonumber
    \\ 
    &\le
    2\nu \E|v^\top x|^\alpha
    +2\E\left[|v^\top x|^\alpha|x^\top\Delta|^\alpha\right]
    && \text{(tower property, noise moment)}
    \\
    &\le
    2\nu
    +2
    \left(\E|v^\top x|^{2\alpha}\right)^{1/2}
    \left(\E|x^\top\Delta|^{2\alpha}\right)^{1/2}
    && \text{($\|v\|_\Sigma=1$, $\alpha\le2$, Cauchy--Schwarz)} \\
    &\le
    2\nu
    +2K_4^{\alpha/2}\|\Delta\|_\Sigma^\alpha
    && \text{($2\alpha \le 4$, definition of $K_4$)}
    \label{eq:mixed-alpha-moment}
\end{align}

Let $R=|y-x^\top\bar\theta|$.
By the definition of $\widehat\tau$ in the lemma statement, if
$\widehat\tau<\inf\{t:\Pr(R\le t)\ge1-128\gamma\}$, then at least
$(1-64\gamma)N$ threshold-block residuals are smaller than this population
quantile, whose strict lower tail is at most $1-128\gamma$.  Hence
\[
    \Pr\left(\widehat\tau<\inf\{t:\Pr(R\le t)\ge1-128\gamma\}\right)
    \le
    \exp(-16N\gamma)
    \le
    \delta/2.
    \qquad\text{(Chernoff)}
\]
Moreover, Markov's inequality gives
$\Pr(R>(\E R^\alpha/(16\gamma))^{1/\alpha})\le16\gamma$.  If
$\widehat\tau>(\E R^\alpha/(16\gamma))^{1/\alpha}$, then more than
$64\gamma N$ threshold-block residuals exceed this deterministic level; it follows that
\[
    \Pr\left(
    \widehat\tau>
    \left(\frac{\E R^\alpha}{16\gamma}\right)^{1/\alpha}
    \right)
    \le
    \exp(-16N\gamma)
    \le
    \delta/2.
    \qquad\text{(Chernoff)}
\]
Hence, with probability at least $1-\delta$ over the threshold block,
\begin{equation}
\label{eq:threshold-event}
    \Pr(|y-x^\top\bar\theta|>\widehat\tau)\le128\gamma,
    \qquad
    \widehat\tau^\alpha
    \le
    \frac{\E|y-x^\top\bar\theta|^\alpha}{16\gamma}.
\end{equation}

We henceforth condition on the threshold block and assume \eqref{eq:threshold-event}.  Define
\[
    g=\Sigma^{-1}\E\left[x(y-x^\top\bar\theta)
    \mathbf 1\{|y-x^\top\bar\theta|\le\widehat\tau\}\right].
\]
The identity $\E[x(y-x^\top\bar\theta)]=\Sigma(\theta^*-\bar\theta)$ gives
\[
    g-(\theta^*-\bar\theta)
    =
    -\Sigma^{-1}\E\left[x(y-x^\top\bar\theta)
    \mathbf 1\{|y-x^\top\bar\theta|>\widehat\tau\}\right].
\]
Therefore,
\begin{align}
    &\|g-(\theta^*-\bar\theta)\|_\Sigma \nonumber\\
    &=\sup_{\|v\|_\Sigma=1}
    \left|\E\left[v^\top x(y-x^\top\bar\theta)
    \mathbf 1\{|y-x^\top\bar\theta|>\widehat\tau\}\right]\right|
    \nonumber\\
    &\le\sup_{\|v\|_\Sigma=1}
    \left(\E\left[|v^\top x|^\alpha|y-x^\top\bar\theta|^\alpha\right]\right)^{1/\alpha}
    \Pr(|y-x^\top\bar\theta|>\widehat\tau)^{\frac{\alpha-1}{\alpha}}
    &&\text{(H\"older)}\nonumber\\
    &\le
    \left(2\nu+2K_4^{\alpha/2}\|\Delta\|_\Sigma^\alpha\right)^{1/\alpha}
    (128\gamma)^{\frac{\alpha-1}{\alpha}}
    &&\text{(\pref{eq:mixed-alpha-moment}, \pref{eq:threshold-event})}\nonumber\\
    &\le
    24\left(
        \nu^{1/\alpha}+\sqrt{K_4}\|\Delta\|_\Sigma
    \right)
    \gamma^{\frac{\alpha-1}{\alpha}}.
    &&\text{($1<\alpha\le2$)}
    \label{eq:population-bias}
\end{align}

On the pruning block, the pruned response decomposes as $z=x^\top g+\xi$.
By construction, $\E[x\xi]=0$.  Since $g$ is the population least-squares
coefficient for the pruned response,
\[
    \E[\xi^2]\le\E[z^2].
\]  
By the definition of $z$ in \pref{alg:iterated-pruning}, we have
$z^2\le \widehat\tau^{2-\alpha}|y-x^\top\bar\theta|^\alpha$, and combining this with \pref{eq:threshold-event} and \pref{eq:residual-alpha-moment}, we obtain
\begin{align}
    \E[z^2]
    \le
    \widehat\tau^{2-\alpha}\E|y-x^\top\bar\theta|^\alpha
    \le
    4
    \left(\nu^{1/\alpha}+\|\Delta\|_\Sigma\right)^2
    \gamma^{-\frac{2-\alpha}{\alpha}}.
    \label{eq:pruned-variance}
\end{align}
The finite-variance guarantee then implies, with probability at least $1-\delta$,
\begin{align}
    \|\widehat g-g\|_\Sigma
    &\le
    2
    \left(\nu^{1/\alpha}+\|\Delta\|_\Sigma\right)
    R_{\rm est}(N,\delta)
    \gamma^{-\frac{2-\alpha}{2\alpha}} \nonumber\\
    &\le
    2 \left(\nu^{1/\alpha}+\|\Delta\|_\Sigma\right)
    \gamma^{\frac{\alpha-1}{\alpha}}, && (R_{\rm est}^2 \le \gamma)
    \label{eq:fv-correction}
\end{align}
Moreover, since $K_4 \ge 1$ by definition,
\[
    \nu^{1/\alpha}+\|\Delta\|_\Sigma
    \le
    \left(1+\sqrt{K_4}\frac{\|\Delta\|_\Sigma}{\nu^{1/\alpha}}\right)
    \nu^{1/\alpha}.
\]
Combining \eqref{eq:population-bias} and \eqref{eq:fv-correction} proves
\eqref{eq:one-step-bound}.  The bound in
\eqref{eq:threshold-event} fails with probability at most $\delta$, and the
finite-variance call fails with probability at most $\delta$, for a total of $2\delta$.
\end{proof}

\section{Proof of \pref{thm:df-blackbox-prediction} (Distribution-Free Prediction)} \label{app:proof_prediction}

To simplify notation, we define
\begin{equation}
    \gamma=
    R_{\rm pred}(n,\delta/2)^2+\frac{\log(4/\delta)}{n}. \label{eq:gamma_def}
\end{equation}
If $\gamma>1/128$, the algorithm returns $\widehat h(x)=0$.
By \pref{assum:raw-response-moment},
\begin{align*}
    |x^\top\theta^*|^\alpha
    &=
    |\E[y\mid x]|^\alpha
    \le
    \E[|y|^\alpha\mid x]
    \le
    \nu,\\
    \E_x\left(\widehat h(x)-x^\top\theta^*\right)^2
    &=
    \E_x\left(x^\top\theta^*\right)^2
    \le
    \nu^{2/\alpha}
    \le
    128\,\nu^{2/\alpha}
    \gamma^{\frac{2(\alpha-1)}{\alpha}}.
\end{align*}
This proves \pref{eq:df-prediction-bound} in the case $\gamma>1/128$.
Assume now that $\gamma\le1/128$.  Since $|T|=n$, we have
$n\gamma\ge\log(4/\delta)$.  Define the deterministic thresholds
\[
    \tau_{\rm quant}:=\inf\{t:\Pr(|y|\le t)\ge1-128\gamma\},
    \qquad
    \tau_{\rm moment}:=\left(\frac{\nu}{16\gamma}\right)^{1/\alpha},
\]
noting that $\Pr(|y|<\tau_{\rm quant})\le1-128\gamma$ by definition.  
By the definition of $\widehat\tau$ in \pref{alg:df-blackbox-prediction},
if $\widehat\tau<\tau_{\rm quant}$, then at most $64\gamma n$ observations in
$T$ satisfy $|y_i| \ge \tau_{\rm quant}$.  Combining this with $\Pr(|y|\ge\tau_{\rm quant})\ge128\gamma$, we have
\begin{equation}
    \Pr\left(\widehat\tau<\tau_{\rm quant}\right)
    \le
    \Pr\left(
    \sum_{(x_i,y_i)\in T}\mathbf 1\{|y_i|\ge\tau_{\rm quant}\}
    \le
    \frac12
    \E\sum_{(x_i,y_i)\in T}\mathbf 1\{|y_i|\ge\tau_{\rm quant}\}
    \right). \label{eq:init_tau_hat}
\end{equation}
We further upper bound the right-hand side as follows:
\begin{align}
    \Pr\left(\widehat\tau<\tau_{\rm quant}\right)
    &\le
    \exp\left(
    -\frac18
    \E\sum_{(x_i,y_i)\in T}\mathbf 1\{|y_i|\ge\tau_{\rm quant}\}
    \right)
    && \text{(\pref{eq:init_tau_hat} and Chernoff)}
    \nonumber\\
    &=
    \exp\left(-\frac n8\Pr(|y|\ge\tau_{\rm quant})\right)
    \nonumber\\
    &\le
    \exp(-16n\gamma)
    && \text{($\Pr(|y|\ge \tau_{\rm quant}) \ge 128\gamma$)}
    \nonumber\\
    &\le
    \delta/4. && \text{(choice of $\gamma$)}
    \label{eq:df-lower-quantile-event}
\end{align}
The upper deterministic threshold satisfies
\begin{align*}
    \Pr(|y|>\tau_{\rm moment})
    &\le
    \frac{\E|y|^\alpha}{\tau_{\rm moment}^\alpha}
    && \text{(Markov)}
    \\
    &\le
    16\gamma. && \text{(moment bound, choice of $\tau_{\rm moment}$)}
\end{align*}
Next, using the definition of $\widehat\tau$, we observe that if $\widehat\tau>\tau_{\rm moment}$, then more than $64\gamma n$ observations
in $T$ exceed $\tau_{\rm moment}$.  It follows that
\begin{align}
    \Pr\left(\widehat\tau>\tau_{\rm moment}\right)
    &\le
    \Pr\left(
    \sum_{(x_i,y_i)\in T}\mathbf 1\{|y_i|>\tau_{\rm moment}\}
    >
    64\gamma n
    \right)
    \nonumber\\
    &\le
    \left(
    \frac{e\,\E\sum_{(x_i,y_i)\in T}
    \mathbf 1\{|y_i|>\tau_{\rm moment}\}}
    {64\gamma n}
    \right)^{64\gamma n}
    && \text{(Chernoff)}
    \nonumber\\
    &=
    \left(
    \frac{e\,\Pr(|y|>\tau_{\rm moment})}
    {64\gamma}
    \right)^{64\gamma n}
    \nonumber\\
    &\le
    \left(\frac e4\right)^{64\gamma n}
    && \text{(Markov)}
    \nonumber\\
    &\le
    \exp(-16n\gamma)
    \nonumber\\
    &\le
    \delta/4. && \text{(choice of $\gamma$)}
    \label{eq:df-upper-quantile-event}
\end{align}
Let $\calE_{\rm thresh}:=\{\tau_{\rm quant}\le\widehat\tau\le
\tau_{\rm moment}\}$.  The bounds \pref{eq:df-lower-quantile-event} and
\pref{eq:df-upper-quantile-event} imply
\[
    \Pr(\calE_{\rm thresh}^c)\le\delta/2.
\]
By the definition of $\tau_{\rm quant}$,
$\Pr(|y|>\tau_{\rm quant})\le128\gamma$.  On $\calE_{\rm thresh}$,
\begin{equation}
\label{eq:df-threshold-event}
    \Pr(|y|>\widehat\tau)\le128\gamma,
    \qquad
    \widehat\tau^\alpha\le\frac{\nu}{16\gamma}.
\end{equation}

Note also that on $\calE_{\rm thresh}$, the threshold is fixed before the independent
pruning block is used.  Let $z$ denote the corresponding population pruned
response, and let $g(x)=\E[z\mid x]$.  We then have
\begin{align*}
    |g(x)-x^\top\theta^*|
    &=
    \left|\E\left[ y\,\mathbf 1\{|y|>\widehat\tau\} \mid x \right] \right|
    \\
    &\le
    \left(\E[|y|^\alpha\mid x]\right)^{1/\alpha} \Pr(|y|>\widehat\tau\mid x)^{\frac{\alpha-1}{\alpha}} && \text{(H\"older)}
    \\
    &\le
    \nu^{1/\alpha} \Pr(|y|>\widehat\tau\mid x)^{\frac{\alpha-1}{\alpha}}. && \text{(response moment)}
\end{align*}
Taking the squared average, we obtain
\begin{align}
    \E_x\left(g(x)-x^\top\theta^*\right)^2
    &\le
    \nu^{2/\alpha}
    \E_x\left[
    \Pr(|y|>\widehat\tau\mid x)^{
    \frac{2(\alpha-1)}{\alpha}}
    \right]
    \nonumber\\
    &\le
    \nu^{2/\alpha}
    \Pr(|y|>\widehat\tau)^{\frac{2(\alpha-1)}{\alpha}}
    && \text{(Jensen and $2(\alpha-1)/\alpha\le1$)}
    \nonumber\\
    &\le
    128\,\nu^{2/\alpha}
    \gamma^{\frac{2(\alpha-1)}{\alpha}}. && \text{(by \pref{eq:df-threshold-event})}
    \label{eq:df-pruning-bias}
\end{align}
The pruned response satisfies
\begin{align*}
    \E[z^2\mid x]
    &\le
    \widehat\tau^{2-\alpha}\E[|y|^\alpha\mid x]
    \le
    \nu\widehat\tau^{2-\alpha}.
\end{align*}
Since the pruning block is independent of the threshold block,
\pref{assum:finite-variance-prediction} gives, with probability at least
$1-\delta/2$ over the pruning block,
\begin{align}
    \E_x\left(\widetilde h(x)-g(x)\right)^2
    &\le
    \E_x\left(x^\top\theta^*-g(x)\right)^2
    +
    \nu\widehat\tau^{2-\alpha}R_{\rm pred}(n,\delta/2)^2
    \nonumber\\
    &\le
    \E_x\left(x^\top\theta^*-g(x)\right)^2
    +
    \nu\widehat\tau^{2-\alpha}\gamma && \text{(choice of $\gamma$)}
    \nonumber\\
    &\le
    129\,\nu^{2/\alpha}
    \gamma^{\frac{2(\alpha-1)}{\alpha}}
    && \text{(\pref{eq:df-pruning-bias} and \pref{eq:df-threshold-event})}
    \label{eq:df-finite-variance-step}
\end{align}

Since $|z|\le\widehat\tau$, we also have $g(x)\in[-\widehat\tau,\widehat\tau]$.
Clipping $\widetilde h$ to this interval cannot increase its squared error to
$g$, so
\pref{eq:df-finite-variance-step} holds with $\widehat h$ in place of
$\widetilde h$.  Combining \pref{eq:df-pruning-bias} and
\pref{eq:df-finite-variance-step},
\begin{align*}
    \E_x\left(\widehat h(x)-x^\top\theta^*\right)^2
    &\le
    2\E_x\left(\widehat h(x)-g(x)\right)^2
    +
    2\E_x\left(g(x)-x^\top\theta^*\right)^2
    \\
    &\le
    514\,\nu^{2/\alpha}
    \gamma^{\frac{2(\alpha-1)}{\alpha}}.
\end{align*}
The threshold event fails with probability at most $\delta/2$, and the
finite-variance prediction call in \pref{eq:df-finite-variance-step} fails
with probability at most $\delta/2$ conditionally on the threshold block.
By the union bound, we conclude that \pref{eq:df-prediction-bound} holds  with probability at least $1-\delta$.

\section{Proofs of Corollaries} \label{sec:proofs_corollaries}

\subsection{Proof of \pref{cor:hanchi-plug-in} (Fourth-Moment Plug-in)}
\begin{proof}[Proof of \pref{cor:hanchi-plug-in}]
The assumed finite-variance rate gives
\[
    R_{\rm est}\left(\frac{n}{8\log n},\frac{\delta}{8\log n}\right)^2
    \le
    32C_{\rm est}^2
    \frac{\log n\left(d+\log(\log n/\delta)\right)}{n}.
\]
The additional term in \eqref{eq:generic-contraction-condition} is bounded by
the same expression with $C_{\rm est}^2$ replaced by $1$.  Choose
$A_{\alpha,C_{\rm est}}=2^{24}(C_{\rm est}^2+1)
\left[A_0+A_0^{\alpha/(\alpha-1)}\right]$.  The assumed lower bound on $n$
implies \eqref{eq:generic-warm-start-sample} and
\[
    R_{\rm est}\left(\frac{n}{8\log n},\frac{\delta}{8\log n}\right)^2
    +
    \frac{\log n\,\log(\log n/\delta)}{n}
    \le
    \frac{2^{19}(C_{\rm est}^2+1)}
    {A_{\alpha,C_{\rm est}}
    K_4^{\,2\vee\frac{\alpha}{2(\alpha-1)}}}.
\]
By the choice of $A_{\alpha,C_{\rm est}}$ and $K_4\ge1$, this is at most 
\[
    \left(\frac{1}{A_0\sqrt{K_4}}\right)^{\frac{\alpha}{\alpha-1}},
\]
which gives \pref{eq:generic-contraction-condition}.

In \citep[Theorem~11]{el2024minimax}, replacing $\delta$ by $\eta$ only changes
the sample-size lower bound through $\log(1/\eta)$.  Using the second term in the assumed lower bound on $n$, together with the preceding control of the logarithmic factors, we have
\[
    \frac{n}{8\log n}
    \ge
    N_{\rm EH}\left(d,\frac{\delta}{8\log n},\calD_X\right),
    \qquad
    \frac n4
    \ge
    N_{\rm EH}\left(d,\frac{\delta}{8},\calD_X\right).
\]
Hence, every block used by \pref{alg:iterated-pruning} is large enough to apply \citep[Theorem~11]{el2024minimax} at the corresponding confidence level.  On the final
block,

\begin{equation}
\label{eq:plugin-final-rate}
    R_{\rm est}\left(\frac{n}{4},\frac{\delta}{8}\right)^2
    +
    \frac{\log(1/\delta)}{n}
    \le
    8(C_{\rm est}^2+1)\frac{d+\log(1/\delta)}{n}.
\end{equation}
Substituting \pref{eq:plugin-final-rate} into
\pref{eq:generic-estimation-bound} gives \pref{eq:hanchi-plugin-bound} with
$B_{\alpha,C_{\rm est}}=
B_0\bigl(8(C_{\rm est}^2+1)\bigr)^{(\alpha-1)/\alpha}$.
\end{proof}

\subsection{Proof of \pref{cor:df-blackbox-prediction} (Distribution-Free Weak-Moment Prediction)}
\begin{proof}[Proof of \pref{cor:df-blackbox-prediction}]
The assumed prediction rate gives 
\begin{align*}
    R_{\rm pred}(n,\delta/2)^2+\frac{\log(4/\delta)}{n}
    &\le
    C_{\rm df}
    \frac{d\log(n/d)+\log(2/\delta)}{n}
    +
    \frac{\log(4/\delta)}{n}
    \\
    &\le
    (C_{\rm df}+1)
    \frac{d\log(n/d)+\log(4/\delta)}{n}.
\end{align*}
Substituting this into \pref{eq:df-prediction-bound} gives 
\pref{eq:df-blackbox-corollary-bound}.
\end{proof}

\section{Mean Estimation with Unknown $\alpha$}
\label{app:empirical-mean-pruning}

We first show that an empirical pruning threshold, estimated from $n$ additional samples, yields the high-probability guarantees needed for the subsequent mean estimation procedure.  This serves as an empirical counterpart to a deterministic pruning threshold used in \cite{cherapanamjeri2022optimal}.

\begin{lemma}[Empirical pruning counterpart of Lemma C.7 in \cite{cherapanamjeri2022optimal}]
\label{lem:empirical-mean-pruning}
Let $1<\alpha<2$, and let $X$ be a mean-zero random vector in
$\mathbb R^d$ such that
\begin{equation}
    \sup_{\|v\|_2=1}\E|\langle X,v\rangle|^\alpha\le1. \label{eq:X_condition}
\end{equation}
Let $\delta\in(0,1)$, and set
\[
    q=\min\left\{
    \frac{1}{64},\,
    \frac{6(d+\log(16/\delta))}{n}
    \right\}.
\]
Let $x^\dagger$ be independent of the threshold and pruning samples (denoted $T$ and $X_1,\dotsc,X_n$ below) and satisfy
$\|x^\dagger\|_2\le24\sqrt d$.  From a threshold block $T$ of size
$n$, 
define
\[
    \widehat\tau
    =
    \max\left\{\sqrt d,\,
    \inf\left\{
    t\ge0:
    \frac1n\sum_{X_i\in T}
    \mathbf 1\{\|X_i-x^\dagger\|_2\le t\}
    \ge1-q
    \right\}
    \right\}.
\]
Let $X_1,\dots,X_n$ be an independent pruning block, set
$\calY=\{X_i:\|X_i-x^\dagger\|_2\le\widehat\tau\}$, and let $Y$ have the
conditional law of $X$ given $\|X-x^\dagger\|_2\le\widehat\tau$.  Denote the
mean and covariance of $Y$ by $\widetilde\mu$ and $\widetilde\Sigma$.  With
probability at least $1-\delta/8-3e^{-n/1000}$, the following hold:
\[
\begin{array}{ll}
\textnormal{(i)}\quad |\calY|\ge 3n/4,
&
\textnormal{(ii)}\quad
\|\widetilde\mu\|_2
\le
2q^{\frac{\alpha-1}{\alpha}},
\\[3mm]
\textnormal{(iii)}\quad
\displaystyle
\sup_{\|v\|_2=1}
\E|\langle Y-\widetilde\mu,v\rangle|^\alpha
\le20,
&
\textnormal{(iv)}\quad
\tr(\widetilde\Sigma)
\le
20d\,q^{-\frac{2-\alpha}{\alpha}}.
\end{array}
\]
\end{lemma}

\begin{proof}
For $u\in(0,1)$, let
$\tau(u)=\inf\{t:\Pr(\|X-x^\dagger\|_2\le t)\ge1-u\}$.  Since
$\Pr(\|X-x^\dagger\|_2<\tau(2q))\le1-2q$,
\[
\begin{aligned}
    \Pr\left(\widehat\tau<\max\{\sqrt d,\tau(2q)\}\right)
    &\le
    \Pr\left(
    \frac1n\sum_{X_i\in T}
    \mathbf 1\{\|X_i-x^\dagger\|_2\ge\tau(2q)\}
    \le q
    \right)
    \\
    &\le e^{-nq/4}.
    && \text{(Chernoff)}
\end{aligned}
\]
Moreover, $\Pr(\|X-x^\dagger\|_2>\tau(q/2))\le q/2$, so
\[
\begin{aligned}
    \Pr\left(\widehat\tau>\max\{\sqrt d,\tau(q/2)\}\right)
    &\le
    \Pr\left(
    \frac1n\sum_{X_i\in T}
    \mathbf 1\{\|X_i-x^\dagger\|_2>\tau(q/2)\}
    >q
    \right)
    \\
    &\le e^{-nq/6}.
    && \text{(Chernoff)}
\end{aligned}
\]
Let $\calE$ denote the event that both displayed threshold bounds hold.  Under $\calE$, the definition of $\tau(2q)$ gives
\begin{equation}
\label{eq:empirical-pruning-tail}
    \Pr(\|X-x^\dagger\|_2>\widehat\tau)\le2q.
\end{equation}
Moreover, Markov's inequality gives
\begin{equation}
    \Pr\left(
    \|X\|_2>
    \left(\frac{2}{q}\E\|X\|_2^\alpha\right)^{1/\alpha}
    \right)
    \le\frac q2. \label{eq:Markov_step}
\end{equation}
We also make use of the following moment upper bound from
\cite[Lemma 4.1]{cherapanamjeri2022optimal}:
\begin{equation}
    \E\|X\|_2^\alpha
    \le
    \frac43 d^{\alpha/2}. \label{eq:moment_step}
\end{equation}
Combining the above, we have
\[
\begin{aligned}
    \tau(q/2)
    &\le
    \|x^\dagger\|_2 + \left(\frac{2}{q}\E\|X\|_2^\alpha\right)^{1/\alpha}
    && \textnormal{(definition of $\tau$, triangle inequality, \pref{eq:Markov_step})}
    \\
    &\le
    24\sqrt d+
    \left(\frac{8}{3q}d^{\alpha/2}\right)^{1/\alpha}
    && \textnormal{($\|x^\dagger\|_2\le24\sqrt d$, \pref{eq:moment_step})}
    \\
    &\le
    6\sqrt d\,q^{-1/\alpha}.
    && \textnormal{($q \le 1/64$, $1 < \alpha < 2$)}
\end{aligned}
\]
Combining with $\widehat\tau \le \max\{\sqrt{d},\tau(q/2)\}$ therefore gives
\begin{equation}
\label{eq:empirical-pruning-radius}
    \widehat\tau
    \le
    6\sqrt d\,q^{-1/\alpha}.
\end{equation}

On $\calE$, each pruning-block point is retained with probability at least
$1-2q\ge31/32$.  Therefore
\[
\begin{aligned}
    \Pr\left(|\calY|<\frac{3n}{4}\right)
    &\le
    \exp\left(-2n\left(\frac{31}{32}-\frac34\right)^2\right)
    && \text{(Hoeffding)}
    \\
    &\le e^{-n/11}.
\end{aligned}
\]

It remains to verify the three population claims.  Since
\pref{eq:empirical-pruning-tail} and $q \le 1/64$ imply
$\Pr(\|X-x^\dagger\|_2\le\widehat\tau)\ge31/32$, we have for any $\|v\|_2=1$ that
\[
\begin{aligned}
    |\langle\widetilde\mu,v\rangle|
    &=
    \frac{
    \left|\E[\langle X,v\rangle
    \mathbf 1\{\|X-x^\dagger\|_2>\widehat\tau\}]\right|
    }{
    \Pr(\|X-x^\dagger\|_2\le\widehat\tau)
    } && \textnormal{(definition of $\widetilde\mu$)}
    \\
    &\le
    \frac{32}{31}
    \Pr(\|X-x^\dagger\|_2>\widehat\tau)^{\frac{\alpha-1}{\alpha}}
    && \text{(Hölder and $\Pr(\cdot) \ge 31/32$)}
    \\
    &\le
    2q^{\frac{\alpha-1}{\alpha}}. 
    && \text{(by \pref{eq:empirical-pruning-tail})}
\end{aligned}
\]
Taking the supremum over $\|v\|_2=1$ gives the bound on
$\|\widetilde\mu\|_2$.  The same $31/32$ lower bound on the retained probability also gives
\begin{align*}
    \E|\langle Y-\widetilde\mu,v\rangle|^\alpha
    &\le
    2^{\alpha-1}\E|\langle Y,v\rangle|^\alpha
    +
    2^{\alpha-1}|\langle\widetilde\mu,v\rangle|^\alpha
    && \text{($|a-b|^\alpha\le2^{\alpha-1}(|a|^\alpha+|b|^\alpha)$)}
    \\
    &\le
    2^{\alpha-1}\frac{32}{31}
    +
    2^{\alpha-1}\|\widetilde\mu\|_2^\alpha && \text{(conditioning defining $Y$)}
    &&
    \\
    &\le20.
    && \text{($\|\widetilde\mu\|_2$ bound and $q\le1/64$)}
\end{align*}

Finally, we have
\[
\begin{aligned}
    \tr(\widetilde\Sigma)
    &\le
    \E\|Y\|_2^2
    \\
    &\le
    \frac{32}{31}\E\left[
    \|X\|_2^2
    \mathbf 1\{\|X-x^\dagger\|_2\le\widehat\tau\}
    \right] && \text{(conditioning defining $Y$)}
    && 
    \\
    &\le
    \frac{32}{31}\E\left[
    \|X\|_2^2
    \mathbf 1\{\|X\|_2\le\widehat\tau+24\sqrt d\}
    \right]
    && \text{($\|x^\dagger\|_2\le24\sqrt d$)}
    \\
    &\le
    \frac{32}{31}
    (\widehat\tau+24\sqrt d)^{2-\alpha}
    \E\|X\|_2^\alpha
    && \text{($u^2\mathbf 1\{u\le R\}\le R^{2-\alpha}u^\alpha$)}
    \\
    &\le
    \frac{32}{31}\cdot\frac43 d^{\alpha/2}
    (\widehat\tau+24\sqrt d)^{2-\alpha} && \text{(by \pref{eq:moment_step})}
    && 
    \\
    &\le
    20d\,q^{-\frac{2-\alpha}{\alpha}}.
    && \text{(by \pref{eq:empirical-pruning-radius})}
\end{aligned}
\]

Recall that $q=\min\big\{\frac{1}{64},\,\frac{6(d+\log(16/\delta))}{n}\big\}$.  
If $q=6(d+\log(16/\delta))/n$, then $e^{-nq/6}\le\delta/16$.  If
$q=1/64$, then $e^{-nq/6}\le e^{-n/1000}$.  The same bound controls
$e^{-nq/4}$, and $e^{-n/11}\le e^{-n/1000}$.  This proves all four claimed
properties with failure probability at most $\delta/8+3e^{-n/1000}$.
\end{proof}

\begin{corollary}[Empirically tuned weak-moment mean estimation]
\label{cor:empirical-mean-estimation}
Let $1<\alpha<2$, and let $X$ have mean $\mu\in\mathbb R^d$ and satisfy
\[
    \sup_{\|v\|_2=1}
    \E|\langle X-\mu,v\rangle|^\alpha\le1.
\]
Assume $\delta\in(0,1/2)$ and $\delta\ge64e^{-n/1000}$.  There is a
modification of the mean estimation algorithm of
\citet{cherapanamjeri2022optimal} that does not use $\alpha$ in the pruning
step and satisfies, with probability at least $1-\delta$,
\[
    \|\widehat\mu-\mu\|_2
    \le
    C
    \left[
    \sqrt{\frac dn}
    +
    \left(\frac{d+\log(1/\delta)}{n}\right)^{
    \frac{\alpha-1}{\alpha}}
    \right],
\]
where $C>0$ is a numerical constant.  The modification uses an independent
threshold block of size $n$ to compute the empirical radius in
\pref{lem:empirical-mean-pruning}, then replaces the deterministic pruning
radius in \citet[Algorithm~5]{cherapanamjeri2022optimal} by this empirical
radius.
\end{corollary}

\begin{proof}
The initial estimate is computed as in
\citet[Algorithm~4]{cherapanamjeri2022optimal}.
\citep[Lemma~B.1]{cherapanamjeri2022optimal} gives
\[
    \|x^\dagger-\mu\|_2\le24\sqrt d
\]
with probability at least $1-e^{-n/50}$.  On this event,
\pref{lem:empirical-mean-pruning}, applied to $X-\mu$ with failure probability
$\delta/2$, gives the four pruning properties used in
\citet[Lemma~C.7]{cherapanamjeri2022optimal}: at least $3n/4$ retained samples, pruning
bias at most $2q^{(\alpha-1)/\alpha}$, retained directional
$\alpha$-moments bounded by $20$, and retained trace covariance at most
$20d q^{-(2-\alpha)/\alpha}$, where
$q=\min\{1/64,6(d+\log(32/\delta))/n\}$.

The rest of their proof uses \citet[Lemma~C.7]{cherapanamjeri2022optimal}
only through these four properties.
Replacing their deterministic pruning radius by the empirical radius from
\pref{lem:empirical-mean-pruning} therefore leaves the testing and descent
steps unchanged, and yields
\[
    \|\widehat\mu-\mu\|_2
    \le
    C
    \left[
    \sqrt{\frac dn}
    +
    \left(\frac{d+\log(1/\delta)}{n}\right)^{
    \frac{\alpha-1}{\alpha}}
    \right].
\]
The only additional cost is the threshold-block failure probability in
\pref{lem:empirical-mean-pruning}.  The condition
$\delta\ge64e^{-n/1000}$ absorbs this term, together with the initialization
failure, into the final failure probability $\delta$.
\end{proof}

\section{Proof of \pref{thm:random-design-lower} (Lower Bound)} \label{app:proof_lower}
\begin{proof} 
Let
\[
    \Delta=\frac14\nu^{1/\alpha}d^{-1/2}
    \left(\frac dn\right)^{\frac{\alpha-1}{\alpha}}
\]
and set
\[
    \Theta=\{\theta\in\bbR^d:(\Sigma^{1/2}\theta)_j\in\{\pm\Delta\}
    \text{ for every }j\in[d]\}.
\]

For each $\theta\in\Theta$, define $\calD_\theta$ as follows.  The covariate is
uniform on the $2d$ points
\[
    \left\{\pm\sqrt d\,\Sigma^{1/2}e_j:j\in[d]\right\}.
\]
Conditionally on $x$, define the response by
\[
    y =
    \begin{cases}
        \displaystyle \frac{4n}{d}x^\top\theta,
        &\displaystyle\text{with probability }\frac{d}{4n},\\[2mm]
        0,&\text{otherwise}.
    \end{cases}
\]
Since $n\ge d$, the probability $d/(4n)$ is at most one.  Let
$P_\theta=\calD_\theta^{\otimes n}$ be the law of the sample under parameter
$\theta$.

We first verify the assumptions.  The covariance is
\begin{align*}
    \E[xx^\top]
    &=
    \frac{1}{d}\sum_{j=1}^d
    d\,\Sigma^{1/2}e_je_j^\top\Sigma^{1/2}
    =
    \Sigma.
\end{align*}
Moreover, for every $v\in\bbR^d$,
\begin{align*}
    \E[(x^\top v)^4]
    &=
    d\sum_{j=1}^d(\Sigma^{1/2}v)_j^4
    \le
    d\left(v^\top\Sigma v\right)^2.
\end{align*}
Hence, the design satisfies 
\pref{assum:fourth-moment-design} with $K_4\le d$.
For any $\theta\in\Theta$ and any point $x$ in the support of the design, we have
\begin{equation}
\label{eq:lower-mean-magnitude}
    |x^\top\theta|
    =
    \sqrt d\,\Delta
    =
    \frac{1}{4}\nu^{1/\alpha}
    \left(\frac{d}{n}\right)^{\frac{\alpha-1}{\alpha}}.
\end{equation}
Moreover,
\begin{align*}
    \E[y\mid x]
    &=
    \frac{d}{4n}\cdot \frac{4n}{d}x^\top\theta
    =
    x^\top\theta.
\end{align*}
The raw response moment is
\begin{align}
\label{eq:lower-raw-response-moment}
    \E[|y|^\alpha\mid x]
    &=
    \frac{d}{4n}\left(\frac{4n}{d}\right)^\alpha |x^\top\theta|^\alpha
    =
    \frac{\nu}{4}.
\end{align}
This also verifies \pref{assum:raw-response-moment} for the parameter $\theta$.  To bound the centered response moment, we write
\begin{align*}
    |x^\top\theta|^\alpha
    &=
    \frac{\nu}{4^\alpha}\left(\frac dn\right)^{\alpha-1}
    && \text{(by \pref{eq:lower-mean-magnitude})}
    \\
    &\le
    \frac{\nu}{4},
    && \text{($n\ge d,\ \alpha\ge1$)}
\end{align*}
and hence
\begin{align*}
    \E[|y-x^\top\theta|^\alpha\mid x]
    &\le
    2^{\alpha-1}\left(\E[|y|^\alpha\mid x]+|x^\top\theta|^\alpha\right)
    && \text{(triangle inequality)}
    \\
    &\le
    2^{\alpha-2}\nu
    && \text{(\pref{eq:lower-mean-magnitude} and \pref{eq:lower-raw-response-moment})}
    \\
    &\le
    \nu.
\end{align*}

We now show that neighboring parameters are hard to distinguish.  For
$j\in[d]$ and $\theta\in\Theta$, define $\theta^{(j)}\in\Theta$ by flipping only
the $j$-th whitened coordinate:
\[
    (\Sigma^{1/2}\theta^{(j)})_j=-(\Sigma^{1/2}\theta)_j,
    \qquad
    (\Sigma^{1/2}\theta^{(j)})_k=(\Sigma^{1/2}\theta)_k\quad k\ne j.
\]
The one-sample laws under $\theta$ and $\theta^{(j)}$ differ only when $x$
selects coordinate $j$, which has probability $1/d$.  Conditional on that event,
the two reward laws agree at zero and place mass $d/(4n)$ on opposite nonzero
atoms.  Hence,
\begin{align}
\label{eq:lower-neighbor-tv}
    \mathrm{TV}(P_\theta,P_{\theta^{(j)}})
    &\le
    n\,\mathrm{TV}(\calD_\theta,\calD_{\theta^{(j)}})
    && \text{(TV tensorization upper bound)}
    \nonumber\\
    &\le
    n\cdot\frac1d\cdot\frac{d}{4n}
    =
    \frac14.
\end{align}

Consider an arbitrary estimator $\widehat{\theta}$ and set
$A_j=\{(\Sigma^{1/2}\widehat{\theta})_j<0\}$.  For each coordinate $j$, define the
average probability of a wrong sign decision by
\[
    e_j=
    \frac{1}{|\Theta|}\sum_{\theta:(\Sigma^{1/2}\theta)_j=\Delta}P_\theta(A_j)
    +
    \frac{1}{|\Theta|}\sum_{\theta:(\Sigma^{1/2}\theta)_j=-\Delta}
    P_\theta(A_j^c).
\]
Reindexing the second sum in the definition of $e_j$ by the flip
$\theta\mapsto\theta^{(j)}$ gives 
\begin{align}
\label{eq:lower-coordinate-sign-error}
    e_j
    &=
    \frac{1}{|\Theta|}\sum_{\theta:(\Sigma^{1/2}\theta)_j=\Delta}
    \left[P_\theta(A_j)+P_{\theta^{(j)}}(A_j^c)\right]
    \nonumber\\
    &\ge
    \frac{1}{|\Theta|}\sum_{\theta:(\Sigma^{1/2}\theta)_j=\Delta}
    \left[1-\mathrm{TV}(P_\theta,P_{\theta^{(j)}})\right]
    \ge
    \frac38.
    && \text{(by \pref{eq:lower-neighbor-tv})}
\end{align}
Let
\[
    M(\theta)=
    \sum_{j=1}^d
    \left[
        \mathbf 1\{(\Sigma^{1/2}\theta)_j=\Delta\}\mathbf 1_{A_j}
        +
        \mathbf 1\{(\Sigma^{1/2}\theta)_j=-\Delta\}\mathbf 1_{A_j^c}
    \right]
\]
be the number of wrong signs.  Summing
\pref{eq:lower-coordinate-sign-error} over $j$ gives
\begin{equation}
\label{eq:lower-average-sign-errors}
    \frac1{|\Theta|}\sum_{\theta\in\Theta} \E_{P_\theta}M(\theta)
    =
    \sum_{j=1}^d e_j
    \ge
    \frac{3d}{8}.
\end{equation}

If a sign is wrong in coordinate $j$, then the $j$-th coordinate of
$\Sigma^{1/2}(\widehat{\theta}-\theta)$ has absolute value at least $\Delta$.
Hence,
\begin{align}
    \|\widehat{\theta}-\theta\|_\Sigma
    = \|\Sigma^{1/2}(\widehat{\theta}-\theta)\|_2
    &\ge
    \Delta M(\theta)^{1/2}
    \nonumber\\
    &\ge
    \frac{\Delta}{\sqrt d}M(\theta). && \textnormal{($M(\theta)\le d$)} \label{eq:lower-estimation-sign-reduction}
\end{align} 
 The fact that \(0\le M(\theta)\le d\) further implies
  \begin{align*}
    \E_{P_\theta} M(\theta)
    &\le
    \frac d4\left(1- P_\theta\Big(M(\theta)\ge\frac d4\Big)\right)
    +
    d\,P_\theta\Big(M(\theta)\ge\frac d4\Big)
    \\
    &=
    \frac d4 + \frac{3d}{4}\, P_\theta\left(M(\theta)\ge\frac d4\right).
\end{align*}
  Averaging over \(\theta\in\Theta\) and using
  \pref{eq:lower-average-sign-errors}, we obtain
  \[
      \frac{3d}{8}
      \le
      \frac d4+\frac{3d}{4}
      \sup_{\theta\in\Theta}
      P_\theta\left(M(\theta)\ge \frac d4\right),
  \]
  and hence
  \begin{align}
\label{eq:lower-sign-errors-probability}
    \sup_{\theta\in\Theta}
    P_\theta\left(M(\theta)\ge\frac d4\right)
    &\ge
    \frac{3d/8-d/4}{3d/4}
    =
    \frac16.
\end{align}
  On this event, \pref{eq:lower-estimation-sign-reduction} gives
  \[
      \|\widehat{\theta}-\theta\|_\Sigma
      \ge
      \Delta\sqrt{\frac d4}
      =
      \frac18\nu^{1/\alpha}
      \left(\frac dn\right)^{\frac{\alpha-1}{\alpha}}.
  \]
  It follows that
  \[
      \sup_{\theta\in\Theta}
      P_\theta\left(
      \|\widehat{\theta}-\theta\|_\Sigma
      \ge
      \frac{1}{8}\nu^{1/\alpha}
      \left(\frac dn\right)^{\frac{\alpha-1}{\alpha}}
      \right)
      \ge
      \frac16,
  \]
  as claimed.

For the linear prediction bound, let $x'$ be an independent fresh
covariate with the common marginal law $\calD_X$.  Conditionally on the training
samples,
\begin{align*}
    \left(\E_{x'}\left[x'^\top(\widehat{\theta}-\theta)\right]^2\right)^{1/2}
    &=
    \|\widehat{\theta}-\theta\|_\Sigma
    =
    \|\Sigma^{1/2}(\widehat{\theta}-\theta)\|_2.
\end{align*}
Combining this identity with \pref{eq:lower-sign-errors-probability} and
\pref{eq:lower-estimation-sign-reduction} gives
\begin{align*}
    \sup_{\theta\in\Theta}
    P_\theta\left(
    \left(
    \E_{x\sim\calD_X}
    \left(x^\top(\widehat{\theta}-\theta)\right)^2
    \right)^{1/2}
    \ge
    \frac{1}{8}\nu^{1/\alpha}
    \left(\frac dn\right)^{\frac{\alpha-1}{\alpha}}
    \right)
    \ge
    \frac16.
\end{align*}

We now use the same sign argument to obtain the $L_2(\calD_X)$ prediction lower bound for arbitrary
predictors.  For such a predictor $\widehat h$, replace $A_j$ in
\pref{eq:lower-coordinate-sign-error} by
\[
    \left\{
    \widehat h(\sqrt d\,\Sigma^{1/2}e_j)
    <
    \widehat h(-\sqrt d\,\Sigma^{1/2}e_j)
    \right\}.
\]
With this modified sign test, the same pairing and TV bound as in \pref{eq:lower-coordinate-sign-error} gives an average wrong-sign probability of at least $3/8$ for each coordinate.  If this sign decision is wrong under
parameter $\theta$, then the two target values on the pair
$\{\sqrt d\,\Sigma^{1/2}e_j,-\sqrt d\,\Sigma^{1/2}e_j\}$ are ordered
oppositely to the two predicted values, and therefore
\[
    \left|\widehat h(\sqrt d\,\Sigma^{1/2}e_j)
        -(\sqrt d\,\Sigma^{1/2}e_j)^\top\theta\right|
    +
    \left|\widehat h(-\sqrt d\,\Sigma^{1/2}e_j)
        -(-\sqrt d\,\Sigma^{1/2}e_j)^\top\theta\right|
    \ge
    2\sqrt d\,\Delta.
\]
Let $M(\theta)$ again denote the number of wrong signs.  The same argument as that leading to \pref{eq:lower-sign-errors-probability} therefore gives
\[
    \sup_{\theta\in\Theta}\E_{P_\theta}M(\theta)\ge\frac{3d}{8},
    \qquad
    \sup_{\theta\in\Theta}
    P_\theta\left(M(\theta)\ge\frac d4\right)
    \ge
    \frac16.
\]
On the event $\{M(\theta)\ge d/4\}$,
\begin{align*}
    \E_{x\sim\calD_X}
    \left(\widehat h(x)-x^\top\theta\right)^2
    &\ge
    \frac{d}{4}\cdot\frac{1}{2d}\cdot 2d\Delta^2
    =
    \frac{d\Delta^2}{4},
\end{align*}
and it follows that
\begin{align*}
    \sup_{\theta\in\Theta}
    P_\theta\left(
    \E_{x\sim\calD_X}
    \left(\widehat h(x)-x^\top\theta\right)^2
    \ge
    \frac{d\Delta^2}{4}
    \right)
    \ge
    \frac16.
\end{align*}
Finally, substituting the value of $\Delta$ gives
\[
    \frac{d\Delta^2}{4}
    =
    \frac{1}{64}\nu^{2/\alpha}
    \left(\frac dn\right)^{\frac{2(\alpha-1)}{\alpha}}.
\]
Since the predictor was arbitrary, this establishes the claimed prediction lower bound.

\end{proof}

\section{Further Related Work} \label{app:further_related}

Mean estimation of random variables over $\bbR$ for heavy-tailed distributions has been widely studied, and tight PAC bounds and efficient estimators that achieve such bounds are known. When the only assumption on the distribution is the existence of variance $\sigma^2$, the empirical mean is not sample-efficient as it can only guarantee the Chebyshev bound $\sigma \sqrt{\frac{1}{n\delta}}$ with probability $1 - \delta$.
Some widely used estimators that achieve the best estimation bounds up to constants are the following. 

\paragraph{Median of means} partitions the data into blocks and computes their mean, and then gives the median of these means as the estimator (\cite{lugosi2019}).  More formally,
\begin{align*}
    \text{MoM}_{ \{B_1, \ldots, B_k\} \vdash [n]}(X_1, \ldots, X_n) = \text{median}\left(\frac{1}{|B_1|} \sum_{i \in B_1} X_i, \ldots,  \frac{1}{|B_k|} \sum_{i \in B_k} X_i  \right),
\end{align*}
where $|B_i| \ge \lfloor \frac{n}{k} \rfloor$. If $k = \lceil 8 \log (2/\delta) \rceil$ then the estimator achieves the tight PAC bound of $\order(\sigma \sqrt{\frac{\log(1/\delta)}{n}})$. If the variance is infinite and one only assumes that the central $\alpha$-moment ($\alpha \in (1, 2)$) is bounded by some value $\nu$, then median-of-means gets a tight PAC bound of $\order\big(\nu^{\frac{1}{\alpha}} \big(\frac{\log(1/\delta)}{n}\big)^\frac{\alpha - 1}{\alpha} \big)$ with the same $k$. Note that while the confidence bound requires knowing $\sigma$ or $(\alpha, \nu)$, the estimator itself only depends on $\delta$.

\paragraph{Trimmed mean} filters or projects samples outside of a threshold and gives empirical means of the refined samples as the estimator.
An order-optimal variant of trimmed mean (\cite{oliveira2025trimmed}) is achieved when partitioning the samples to two equal halves ($(X_{0,i}, X_{1,i})_{i = 1}^n$) of size $n$, re-arranging the samples of first half to be nondecreasing, then projecting all samples of second half on interval $[X_{0,\epsilon n}, X_{0,(1 - \epsilon)n}]$ where $\epsilon = \frac{16 \log(1/\delta)}{3n}$. This instance of trimmed mean is order-optimal in both finite variance and finite $\alpha$ central moment settings, and in addition is robust to certain adversarial contaminations.

\paragraph{Optimal constants and instance-dependent estimators}
While both MoM and trimmed mean estimators achieve the optimal PAC bound, the constants of their guarantees are large, and their PAC bounds are for worst-case distributions and not distribution dependent. For the former issue, \cite{Catoni2012} proposed a novel estimator that generalizes the empirical mean and achieves tight constants; however, it requires knowing the variance $\sigma^2$ contrary to other estimators. \cite{lee2022optimal} achieved tight constants without requiring prior knowledge of $\sigma^2$.
Instance-dependent optimality was recently characterized in \cite{dang2023optimality}, showing that median-of-means achieves instance-optimal rates up to constants.  

Mean estimation for random variables over $\bbR^d$ has also been widely studied, and comes with significant additional challenges including computational efficiency of the estimator.  When the covariance matrix $\Sigma = \E[(X - \mu)(X - \mu)^\top]$ exists, the tight PAC bound (sub-Gaussian) for $l_2$-distance is 
\begin{align*}
    \|\widehat{\mu} - \mu\|_2 \le \order \left(\sqrt{\frac{\tr(\Sigma)}{n}} + \sqrt{\frac{\lambda_{\max}(\Sigma) \log(1/\delta)}{n}} \right) \qquad \text{ w.p. } 1 - \delta.
\end{align*}
There are many extensions of median-of-means to $\bbR^d$, as the median can be defined in several different ways on $\bbR^d$ (e.g., see \cite{mediansurvay}). If we choose the median to be the geometric median (i.e. $\arg\min_m \sum_i \|m - X_i\|_2$), which is a convex objective and can be computed efficiently, then PAC bound of $\sqrt{\frac{\tr(\Sigma) \log(1/\delta)}{n}}$ is achieved (\cite{minsker2015geometric}). However, this bound has a $\sqrt{\log{1/\delta}}$ gap to the optimal bound. An optimal but computationally inefficient median-of-means variant was proposed in \cite{LugosiMendelson2019}, where the median is defined as 
\begin{align*}
    \argmin_{m \in \bbR^d} \text{radius}\left(\left\{x \in \bbR^d : \sum \boldsymbol{1}\left( \| X_i - x\|\le \| X_i - m \| \right) \ge \frac{n}{2} \right\} \right).
\end{align*}
Another formulation of the above (by \cite{hopkins2020robust}) is the $(0.01, \left( \sqrt{\frac{\tr(\Sigma)}{n}} + \sqrt{\frac{\lambda_{\max}(\Sigma) \log(1/\delta)}{n}}\right)^2 )$-combinatorial center of partition means $(Z_i)_{i = 1}^k$, where $m$ is a $(\epsilon, \lambda)$-combinatorial center if for all unit vectors $v \in \bbR^d$
\begin{align*}
    \sum_i \boldsymbol{1}[ (Z_i - m)^\top v \ge \sqrt{\lambda} ] \le \epsilon k
\end{align*}
\cite{hopkins2020mean} proposed a computationally efficient variant of the estimator by semidefinite relaxations of the optimization objective.

For the mean estimation problem, for distributions satisfying a finite directional $\alpha$-moment bound with
$1<\alpha<2$, \citet{cherapanamjeri2022optimal} give a computationally
efficient mean estimator with matching upper and lower bounds.  In contrast, for linear regression, works adopting weak moment assumptions have mainly focused on the finite-variance regime, typically under fixed or random covariate designs. The seminal work of \cite{audibert2011robust} achieved sub-Gaussian bounds with a $L_2-L_4$ norm equivalence assumption. Later \cite{lugosi2019risk} proposed a median-of-means based estimator that achieved a sub-Gaussian rate without any extra assumption other than existence of the fourth moment. However, these estimators are computationally inefficient. 
\cite{cherapanamjeri2020algorithms} proposed an SDP relaxation of the estimator that achieves optimal sub-Gaussian rates with a polynomial-time complexity but requires stronger assumptions on the covariates. 
\cite{sun2020adaptive} proposed a Huber loss based estimator but with suboptimal rates. Later, \cite{pensia2025robust} improved the Huber-loss based estimation to achieve sub-Gaussian rates if the noise variables are assumed to be symmetric, or i.i.d.~and independent from the covariates. Moreover,~\cite{oliveira2025trimmed} proposed a trimmed mean estimator:  
\begin{align*}
    \hat{\theta}_k = \arg\min_{\theta \in \bbR^d} \max_{\psi \in \bbR^d} \frac{1}{n} \sum_i \text{trim}_k( \ell(\theta^\top x_i - y_i) - \ell(\psi^\top x_i - y_i) ),
\end{align*}
where trimming is done on the interval of $k + 1$ to $n - k$ smallest values. For squared loss $\ell$ and $k = 8 \log(4/\delta)$, it achieves the following estimation rate:
\begin{align*}
    \E \left[|y - \hat{\theta}^\top x|^2\right]^{1/2} \le \order \left(g(\calD_X) \sigma \sqrt{\frac{d + \log(1/\delta)}{n}} \right),
\end{align*}
where $g(\calD_X) = \sup_v \frac{\E\left[|v^\top x|^2\right]^{1/2}}{\E[|v^\top x|]}$. This has a $g(\calD_X)$ term with the lower bound achieved by Gaussian noise. \cite{el2024minimax} studied $\ell_p$-error for $p \ge 2$ and proposed an algorithm that achieves sub-Gaussian rates. Beyond the bounded variance error regime, to our knowledge the only known result for regression is \cite{sun2020adaptive} which is suboptimal even in the finite variance case.

A generalization of the above estimation problems is stochastic optimization, where given a distribution $\calF$ over functions $f: \bbR^d \rightarrow \bbR$, we want to estimate the minimizer of the expected function, i.e. 
$\argmin_w \E_{f \sim \calF} [f(w)]$. For linear regression with any loss $\ell$, we can formulate it as a stochastic optimization problem where function $f_{(x, y)}(w) := \ell(x^\top w - y)$ is drawn when $(x, y) \sim \calD$. \cite{das2024near} proposed clipped SGD, which achieves the best known upper bound of 
\begin{align*}
    \sqrt{\frac{\tr(\Sigma)}{n} + \frac{\sqrt{\tr(\Sigma)\|\Sigma\|_2} \log(\log n/\delta)}{n}}.
\end{align*}
Achieving a sub-Gaussian rate for stochastic optimization remains open.

\section{Numerical Experiment}
\label{app:numerical-experiment}

\begin{figure}[H]
    \centering
    \includegraphics[width=0.6\linewidth]{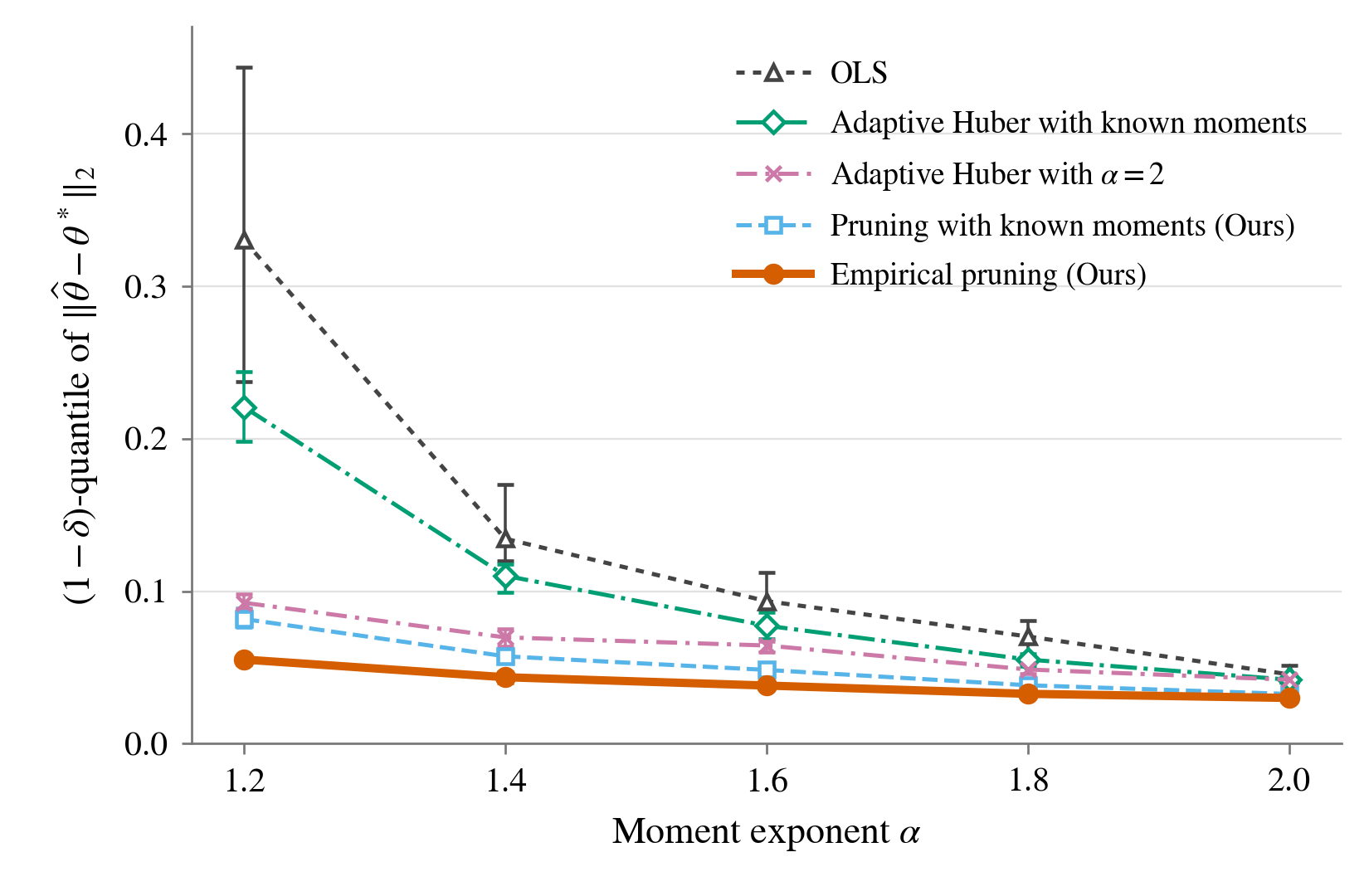}
    \caption{Empirical $(1-\delta)$-quantile of
    $\|\widehat\theta-\theta^*\|_2$, with
    $\delta=1/\log n\approx0.1316$.
    Error bars show $95\%$ bootstrap confidence intervals for each quantile,
    computed using $10{,}000$ bootstrap samples of the $200$ estimation errors.}
    \label{fig:numerical}
\end{figure}

In this section, we present a simple proof-of-concept numerical experiment illustrating
adaptation to unknown $\alpha$ and $\nu$.  We compare the estimation error
of empirical pruning with that of pruning based on known moments as the
noise becomes heavier-tailed.

\paragraph{Setting.}
We consider linear regression with $n=2{,}000$, $d=8$, and
$\theta^*=\mathbf{1}$.  The covariates have independent standardized
Student-$t_5$ coordinates, so $\Sigma=I$ and $K_4=9$.  The noise is independent
of the covariates and follows a centered Pareto distribution with shape
$\alpha+0.1$, rescaled to satisfy $\E|\varepsilon|^\alpha=\nu=1$.
We vary $\alpha\in\{1.2,1.4,1.6,1.8,2\}$ and set $\delta=1/\log n$.
Each experiment is repeated $200$ times, with all methods fitted to the
same data in each run.  We compare our algorithm with OLS and adaptive
Huber regression \citep{sun2020adaptive}, with the latter using either the
true $\alpha$ or $\alpha=2$.

\paragraph{Results.}
In \pref{fig:numerical}, we see that empirical pruning achieves smaller estimation
errors than pruning with known moments throughout the tested range of
$\alpha$.
Both pruning methods have smaller errors than OLS and adaptive Huber
regression, with larger differences as $\alpha$ decreases.

\end{document}